\documentclass[]{pasj02} 

\usepackage{graphicx}
\usepackage{natbib}
\usepackage{multirow}
\usepackage{caption}
\usepackage{url}

\usepackage{amsfonts}

\usepackage{threeparttable}
\jyear{2024}
\Received{}
\Accepted{}

\long\def\comment#1{}
\comment{
   Usage:
     \comment{ comments....}
}

\newcommand{\lw}[1]{\smash{\lower2.ex\hbox{#1}}}
\newcommand{\llw}[1]{\smash{\lower4.ex\hbox{#1}}}

 \definecolor{scarlet}{rgb}{0.7421875, 0.00390625, 0.09765625}
\definecolor{lightgray}{rgb}{0.82422,0.82422,0.82422}
\definecolor{lightgray2}{rgb}{0.75,0.75,0.75}

\usepackage{ulem}
 \DeclareRobustCommand{\erase}{\bgroup\markoverwith{\textcolor{red}{\rule[.5ex]{2pt}{1.6pt}}}\ULon}

\def\erasedtextcolor{lightgray2}
\renewcommand{\erase}[1]{{\leavevmode\textcolor{\erasedtextcolor}{#1}}}

\begin{document} 

\title{Moving object detection from multi-depth images with an attention-enhanced CNN}

\author{
Masato \textsc{Shibukawa},\altaffilmark{1}\orcid{0009-0004-0645-4246} \email{shibukawa.masato@ac.jaxa.jp} 
Fumi \textsc{Yoshida},\altaffilmark{2,3}\orcid{0000-0002-3286-911X}
Toshifumi \textsc{Yanagisawa},\altaffilmark{4}\orcid{0009-0002-7579-9972}
Takashi \textsc{Ito},\altaffilmark{3,5,6}\orcid{0000-0002-0549-9002}\\
Hirohisa \textsc{Kurosaki},\altaffilmark{7}\orcid{0009-0005-3585-0771}
Makoto \textsc{Yoshikawa},\altaffilmark{8}\orcid{0000-0002-3118-7475}
Kohki \textsc{Kamiya},\altaffilmark{7}\orcid{0000-0002-3517-4751}
Ji-an \textsc{Jiang},\altaffilmark{9,10}\orcid{0000-0002-9092-0593}
Wesley \textsc{Fraser},\altaffilmark{11,12}\orcid{0000-0001-6680-6558}
JJ \textsc{Kavelaars},\altaffilmark{11,12}\orcid{0000-0001-7032-5255}
Susan \textsc{Benecchi},\altaffilmark{13}\orcid{0000-0001-8821-5927}
Anne \textsc{Verbiscer},\altaffilmark{14,15}\orcid{0000-0002-3323-9304}
Akira \textsc{Hatakeyama},\altaffilmark{1}\orcid{0009-0008-7516-5543}
Hosei \textsc{O},\altaffilmark{16}\orcid{0009-0007-4828-7185}\\
Naoya \textsc{Ozaki},\altaffilmark{1,8}\orcid{0000-0002-8445-1575}
}

\altaffiltext{1}{The Graduate University for Advanced Studies, SOKENDAI, Shyonan International Village, Hayama, Miura, Kanagawa 240-0193, Japan}
\altaffiltext{2}{University of Occupational and Environmental Health, Japan, 1-1 Iseigaoka, Yahata, Kitakyusyu, Fukuoka 807-8555, Japan}
\altaffiltext{3}{Planetary Exploration Research Center, Chiba Institute of Technology, 2-17-1 Tsudanuma, Narashino, Chiba 275-0016, Japan}
\altaffiltext{4}{Star Signal Solutions Inc, Jindaiji Higashimachi 7-44-1, Chofu, Tokyo 182-8522, Japan}
\altaffiltext{5}{Center for Computational Astrophysics, National Astronomical Observatory of Japan, Osawa 2-21-1, Mitaka, Tokyo 181-8588, Japan}
\altaffiltext{6}{College of Science and Engineering, Chubu University, 1200 Matsumoto-cho, Kasugai, Aichi 487-8501, Japan}
\altaffiltext{7}{Chofu Headquarters, Japan Aerospace Exploration Agency, Jindaiji Higashimachi 7-44-1, Chofu, Tokyo 182-0012, Japan}
\altaffiltext{8}{ISAS, Japan Aerospace Exploration Agency, Yoshinodai 3-1-1, Sagamihara, Kanagawa 252-0222, Japan}
\altaffiltext{9}{Department of Astronomy, University of Science and Technology of China, Hefei 230026, China}
\altaffiltext{10}{Division of Science, National Astronomical Observatory of Japan, Osawa 2-21-1, Mitaka, Tokyo 181-8588, Japan}
\altaffiltext{11}{National Research Council of Canada, Herzberg Astronomy and Astrophysics Research Centre, 5071 W. Saanich Rd., Victoria, BC, V9E 2E7, Canada}
\altaffiltext{12}{Department of Physics and Astronomy, University of Victoria, Elliott Building, 3800 Finnerty Road, Victoria, BC, V8P 5C2, Canada}
\altaffiltext{13}{Planetary Science Institute, 1700 East Fort Lowell, Suite 106, Tucson, AZ 85719, USA}
\altaffiltext{14}{Southwest Research Institute, 1050 Walnut Street, Boulder, CO 80302, USA}
\altaffiltext{15}{Department of Astronomy, University of Virginia, P.O. Box 400325, Charlottesville, VA 22904-4325, USA}
\altaffiltext{16}{University of Tokyo, 7-3-1 Hongo, Bunkyo, Tokyo 113-8654, Japan}



\maketitle

\begin{abstract}
\clearpage
One of the greatest challenges for detecting moving objects in the solar system from wide-field survey data is determining whether a signal indicates a true object or is due to some other source, like noise.
Object verification has relied heavily on human eyes, which usually results in significant labor costs.
In order to address this limitation and reduce the reliance on manual intervention, we propose a multi-input convolutional neural network integrated with a convolutional block attention module.
This method is specifically tailored to enhance the moving object detection system that we have developed and used previously.
The current method introduces two innovations.
This first one is a multi-input architecture that processes multiple stacked images simultaneously.
The second is the incorporation of the convolutional block attention module which enables the model to focus on essential features in both spatial and channel dimensions.
These advancements facilitate efficient learning from multiple inputs, leading to more robust detection of moving objects.
The performance of the model is evaluated on a dataset consisting of approximately 2,000 observational images.
We achieved an accuracy of nearly 99{\%} with AUC (an Area Under the Curve) of $>0.99$.
These metrics indicate that the proposed model achieves excellent classification performance.
By adjusting the threshold for object detection, the new model reduces the human workload by more than {99\%}
compared to manual verification.
\end{abstract}


\section{Introduction\label{sec:intro}}
Trans-Neptunian Objects (TNOs), also known as Kuiper Belt Objects (KBOs), are primordial bodies that retain information from the early Solar System.
Their exploration is crucial for understanding planetary formation theories and the evolution of the Solar System.
NASA's New Horizons (hereafter referred to as NH) mission was launched with the objective of conducting flyby explorations of Pluto and other KBOs, successfully achieving close observations of Pluto in 2015 and Arrokoth in 2019 \citep{stern2015,stern2019}.

Afterwards, the spacecraft began navigating beyond the studied Kuiper Belt region to observe KBOs at resolutions and phase angles unavailable from Earth orbit.
Our team continued to search for observable KBOs while the spacecraft traveled outward \citep{porter2016,verbiscer2019,verbiscer2022}.

In May 2020, Japanese scientists officially joined the New Horizons (NH) ground-based campaign.
Using the Hyper Suprime-Cam (HSC) at the Subaru Telescope, Mauna Kea, Hawaii \citep{miyazaki2018}, the team has contributed to the discovery of KBOs that are potentially observable from the NH spacecraft \citep{fraser2024,yoshida2024}.

KBOs are extremely faint and are often confused with background stars.
This makes their detection from the ground a persistently challenging task.
To reliably capture such faint and distant objects, an observational strategy is required that involves repeatedly observing the same celestial region and stacking numerous images to extend the effective exposure time and achieve a deep limiting magnitude. 
For this purpose, we have relied on a moving detection system \citep{yanagisawa2005,yanagisawa2021} that has been developed by the Japan Aerospace Exploration Agency (hereafter referred to as JAXA).
This system (hereafter referred to as the JAXA system) effectively separates the signals of faint, moving objects like KBOs from background noise and transform them into candidate images suitable for detection.
The JAXA system was originally developed for the observation of small and fast-moving space debris.
However, with the abundance of observational data available, there are always quantitative difficulties associated with this task:
Manually inspecting all of these to identify true celestial objects requires an enormous amount of time and effort, making it impractical.  
This necessitate the ability to automatically detect target objects efficiently and with high precision from a massive volume of data.

Recently the amount of data far exceeds human processing capabilities, which directly motivates the introduction of machine learning techniques (ML).
Most ML-based object detection models in astronomy have operated on single-image inputs.
This choice is partly driven by the simplicity of model architecture and data handling.
However, it overlooks the richness of astronomical observations, which often include time-series and multi-wavelength images of the same sky region.
Integrating such multi-frame data can enhance detection performance by improving signal-to-noise ratios and capturing temporal and spectral features.
In high-contrast imaging, for example, ML pipelines such as Deep PACO \citep{flasseur2024deep} and MODEL{\&}CO \citep{bodrito2024model} have been developed to leverage temporal, spatial, and spectral information.
However, these methods are tailored for stationary sources and are not suitable for moving objects that shift across image frames.
A notable approach for moving object detection is DeepStreaks \citep{duev2019real}, a deep learning model developed for the Zwicky Transient Facility (ZTF) to classify transient events as real or bogus.
It uses a triplet of images—science, reference, and difference—as input.
However, it does not perform image stacking in preprocessing and targets relatively shallow detections, limiting its sensitivity to faint objects.

Compared with preceding studies along this line, our present study address the issues mentioned earlier, and tries to exploit information from images with multiple and stacking depths.
Specifically, we propose a new convolutional neural network(CNN) model that simultaneously processes images ``stacked'' to four different depths (4, 8, 16, and 32 frames) as a three-dimensional tensor.
We further enhance this multi-depth detection model by incorporating the Convolutional Block Attention Module \citep[CBAM;][]{woo2018}, which improves feature learning in both channel and spatial dimensions.  
We validate the effectiveness of the proposed model using actual observational data, and benchmark its performance against various input image combinations and other model architectures.
We assess its potential to reduce the workload of human inspection.

This structure of this article is as follows.
Section~\ref{sec:dataset} describes the dataset used in this study, including its origin, structure, and preprocessing procedures.  
Section~\ref{sec:method} outlines the ML framework employed, along with the evaluation metrics and training strategies.  
In Section~\ref{sec:result}, we present the main experimental results and performance comparisons.  
Finally, Section~\ref{sec:discussion} discusses the implications of the findings and summarizes the key contributions of this study.

\section{Dataset and Image Processing\label{sec:dataset}}
The observations presented herein were conducted as part of the broader NH mission activity.
The details of the observation procedure and the specification of data have been already described in \citep[][their Table 1]{yoshida2024}.
The detail of the current dataset can be found in Table \ref{tab:transposed_data_reversed}.
The data consists of continuous images taken of an area (we call the area ``F2'') corresponding to one HSC field of view.
This field is selected along the direction of travel of the NH spacecraft over the course of half a night.
Note that all the datasets used in this study are publicly accessible via the SMOKA Science Archive (\url{https://smoka.nao.ac.jp/}).

\begin{table}[htbp]
\centering
\caption{Selected and Transposed Observational Datasets (Red. id 03093 \& 03072)}
\label{tab:transposed_data_reversed}
\resizebox{\linewidth}{!}{
\begin{tabular}{|c|c|c|} 
\hline
\textbf{Parameter} & \textbf{Dataset 1} & \textbf{Dataset 2} \\
\hline
DATEOBS (UT) & 2020/06/20 & 2020/05/30 \\
\hline
Red. id & 03093 & 03072 \\
\hline
\begin{tabular}{l}
Start time \\of the first \\exposure
\end{tabular} & 10:55:39.175 & 10:31:55.756 \\
\hline
\begin{tabular}{l}
Start time \\of the last \\exposure 
\end{tabular} & 14:55:24.839 & 14:50:04.288 \\
\hline
RA2000 ( ${}^{\circ}$ ) & 288.662 & 288.714 \\
\hline
DEC2000 ( ${}^{\circ}$ ) & $-20.049$ & $-20.380$ \\
\hline
Field id & F2 & F2 \\
\hline
Filter & $r2$ & $r2$ \\
\hline
Exp time (s) & 90 & 90 \\
\hline
Number of images & 118 & 128 \\
\hline
Night condition & 
\begin{tabular}{c}
(Sky) high clouds, \\
(Seeing) $0^{\prime \prime} 7-1^{\prime \prime} 1$ \\ 
(Temp) $6.9-7.6^{\circ} \mathrm{C}$\\
(Wind) $1.9-7.7 \mathrm{~m} \mathrm{~s}^{-1}$ \\ 
(Humidity) 2.2\%-24.4\% 
\end{tabular}
&
\begin{tabular}{c}
(Sky) clear, \\
(Seeing) 0."7-1".2 \\
(Temp) $3.6-5.9^{\circ} \mathrm{C}$,\\
(Wind) $2.5-15.4 \mathrm{~m} \mathrm{~s}^{-1}$ \\ 
(Humidity) $17.5 \%-28.5 \%$ \\
\end{tabular}\\
\hline
\end{tabular}
}
\end{table}

\begin{figure}[htbp]
  \centering
  \includegraphics[width=\linewidth]{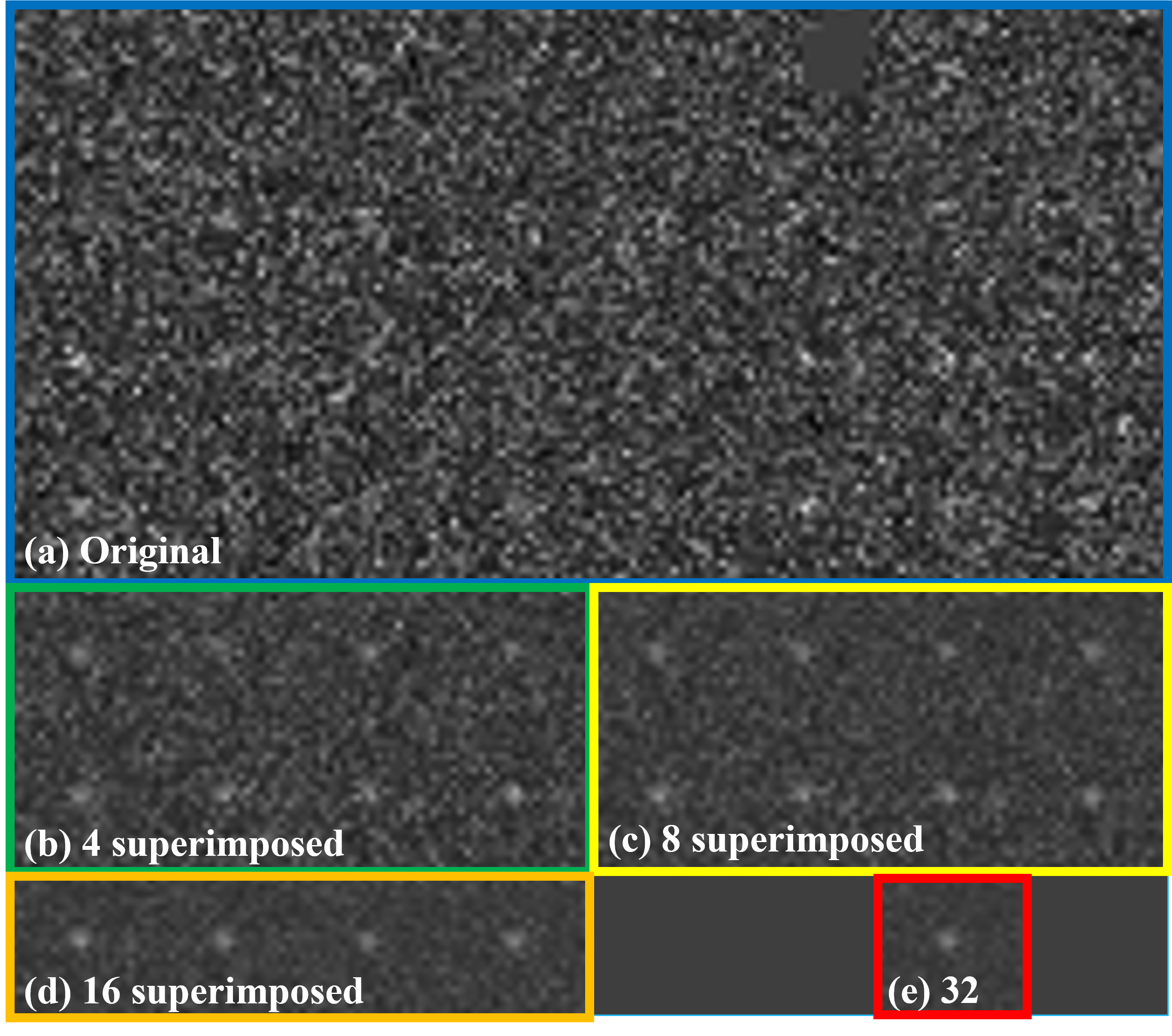}
  \caption{An example of images generated by our moving object detection system.
  (a) Raw observation images. (b) 4-frame stacked images. (c) 8-frame stacked images. (d) 16-frame stacked images. (e) A 32-frame stacked image.
  {Alt text: Figure with five panels showing astronomical cutouts. Panel a is a single noisy raw image. Panels b to e are stacks from 4, 8, 16, and 32 frames; noise decreases and faint sources become progressively visible.}}
  \label{fig:sup_images}
\end{figure}

The JAXA system searches for moving objects using a set of 32 images taken at equal time intervals from the image sets collected each night.
The employed methodology is based on multi-frame image stacking, a technique designed to enhance the signal-to-noise ratio (S/N) of target objects relative to background noise, thereby improving their visibility.
The process begins with the implementation of fundamental noise reduction techniques, including dark-frame subtraction and sky background removal.
Subsequently, multiple frames are extracted from a single field of view, with pixel shifts estimated based on the expected motion of the target object.
These cropped regions are then combined using a median filter.
When the assumed shift aligns with the actual motion of the object, a concentrated bright spot appears at the center of the resulting image.
The system performs stacking with sets of 4, 8, and 16 frames, and finally generates a 32-frame stacked image.
The resulting outputs are the stacked images shown in Figure~\ref{fig:sup_images}.
A single night of observation typically produces approximately 2,000 images, each of which must be manually inspected by a human operator.
During this inspection process, the operator classifies each detected source as either a genuine celestial object or noise.
Because the JAXA moving-object detection system provides both the morphological characteristics of each light source and its apparent motion, human classification is generally reliable.
Nevertheless, the manual assessment of several thousand images per night constitutes a considerable operational burden.

To reduce this workload, a machine-learning–based detection framework has been developed to automatically identify moving objects with high precision and efficiency.
The teacher data were constructed from a highly reliable human classification of the datasets shown in Table~\ref{tab:transposed_data_reversed}.
This system is designed to process large volumes of observational data while maintaining robust performance across diverse observational conditions.

\begin{figure*}[htbp]
  \centering
  \includegraphics[width=\linewidth]{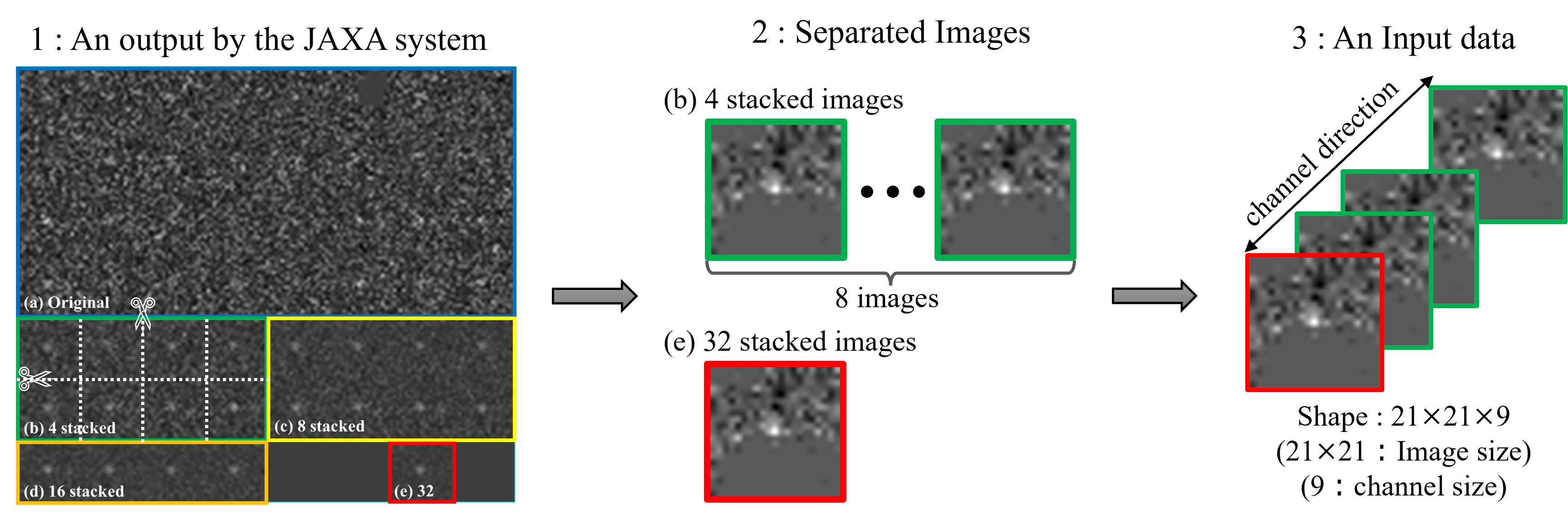}
  \caption{Block diagram of input data construction: (1) output from the JAXA detection system, (2) intermediate stacked images generated from 4, 8, 16, and 32 frames, and (3) input tensor formed by aligning the 4-frame and 32-frame stacked images along the channel axis. {Alt text: Three-step flow diagram. First, outputs from the JAXA detection system are taken as inputs. Second, stacks are produced at 4, 8, 16, and 32 frames. Third, the 4-frame and 32-frame stacks are combined along the channel axis to form the model input tensor.}}
  \label{fig:data_processing}
\end{figure*}

\subsection{Pre-processing for machine learning}
Input data for the ML model are generated by stacking images derived from different numbers of frames along the channel axis.
It is important to note that the term ``stacking'' as used herein differs from conventional astronomical image processing techniques such as ``superposition'' or ``co-adding,'' in which multiple images are overlaid and their pixel intensity values are summed.
In the context of this study, stacking refers exclusively to the operation of aligning separately processed images along the channel dimension, effectively treating them as multi-channel data.
This procedure involves no arithmetic operations on the pixel values.
Figure~\ref{fig:data_processing} illustrates this stacking procedure. 
For instance, an image obtained from 32-frame stacking and another from 16-frame stacking are arranged along the channel axis to form a five-channel input tensor.
The input data for the ML model are constructed by stacking intermediate products obtained during the 32-frame stacking process.
Specifically, they are images generated from 4, 8, and 16
frame stackings.
This multi-resolution stacking approach is designed to effectively capture the characteristic features of objects with varying apparent velocities.
For model evaluation, we constructed two distinct ML datasets derived from the HSC observations corresponding to the two nights described in the previous subsection.
A summary of these datasets is presented in Table~\ref{tab:ml_datasets}.
Dataset 1 is used for training and is augmented sixfold using geometric transformations, including $90^\circ$, $180^\circ$, and $270^\circ$ rotations, as well as horizontal and vertical flipping.
Additionally, during channel-wise stacking, the order of image placement along the channel axis is randomly permuted for each augmented sample to improve the model's generalization capability.
Dataset 1 is partitioned into training, validation, and internal test subsets in a 7:1:2 ratio, respectively.
For evaluation purposes, the unseen Dataset 2 is used without any augmentation.
To improve convergence during training, both datasets are standardized to have zero mean and unit variance.
The presence or absence of objects in the datasets is determined through careful visual inspection by researchers.
This labeling process is performed on candidate objects that were previously extracted by the moving object detection system.

\begin{table}[htbp]
\centering
\begin{threeparttable}
\caption{Summary of Machine Learning Datasets}
\label{tab:ml_datasets}
\begin{tabular}{|l|c|c|}
\hline
\textbf{Feature} & \textbf{Dataset 1} & \textbf{Dataset 2} \\
\hline
Red. id & 03093 & 03072 \\
\hline
Primary Use & Training & Evaluating \\
\hline
Total Stacked Images & 1,966 (11,796) & 3,015 \\
\hline
Images with Objects & 1,476 (8,856) & 2,643 \\
\hline
Images without Objects & 490 (2,940) & 372 \\
\hline
\end{tabular}
\begin{tablenotes}
\footnotesize
\item Numbers in parentheses represent data after sixfold augmentation (e.g., 1,966 raw images became 11,796 after stacking).
\end{tablenotes}
\end{threeparttable}
\end{table}


\section{Machine Learning Method\label{sec:method}}
Previous ML-based moving object detection pipelines, such as DeepStreaks \citep{duev2019real}, have primarily relied on single stocked or difference images for classification.
While this design simplifies data handling and model structure, it limits the ability to capture multi-depth information that is crucial for identifying faint objects.
To address these limitations, we propose a convolutional neural network (CNN)–based model that exploits stacked images with multiple depths as input.
The proposed framework further integrates the Convolutional Block Attention Module (CBAM; \citealt{woo2018}) to enhance feature learning in both channel and spatial dimensions, thereby improving detection sensitivity and overall classification performance compared with previous ML-based approaches.

\subsection{Proposed Method}\label{sssec:3.2}
We constructed a model that integrates Convolutional Block Attention Module(CBAM) with multiple Convolutional Neural Network (CNN) backbones of varying scales.
The input is a 3D tensor consisting of multiple stacked images.
The backbone CNN extracts image features, while CBAM enhances their relevance.
The model outputs a probability between 0 and 1, representing the likelihood of object presence, and classification is performed using a fixed threshold.
We experimented with both small-scale CNNs (2-4 layers) and large-scale ResNet architectures \citep{he2015}.
Previous work has shown that CBAM improves image processing performance when incorporated into ResNet \citep{woo2018}, and this study aims to evaluate whether similar improvements occur in our object detection task.
Given the relatively small and imbalanced dataset in this study, overfitting is a concern.
Therefore, we also examine CBAM's effectiveness with smaller models for comparision.

\subsubsection{Small-scale CNNs}
\begin{figure}[htbp]
  \centering
  \includegraphics[width=\linewidth]{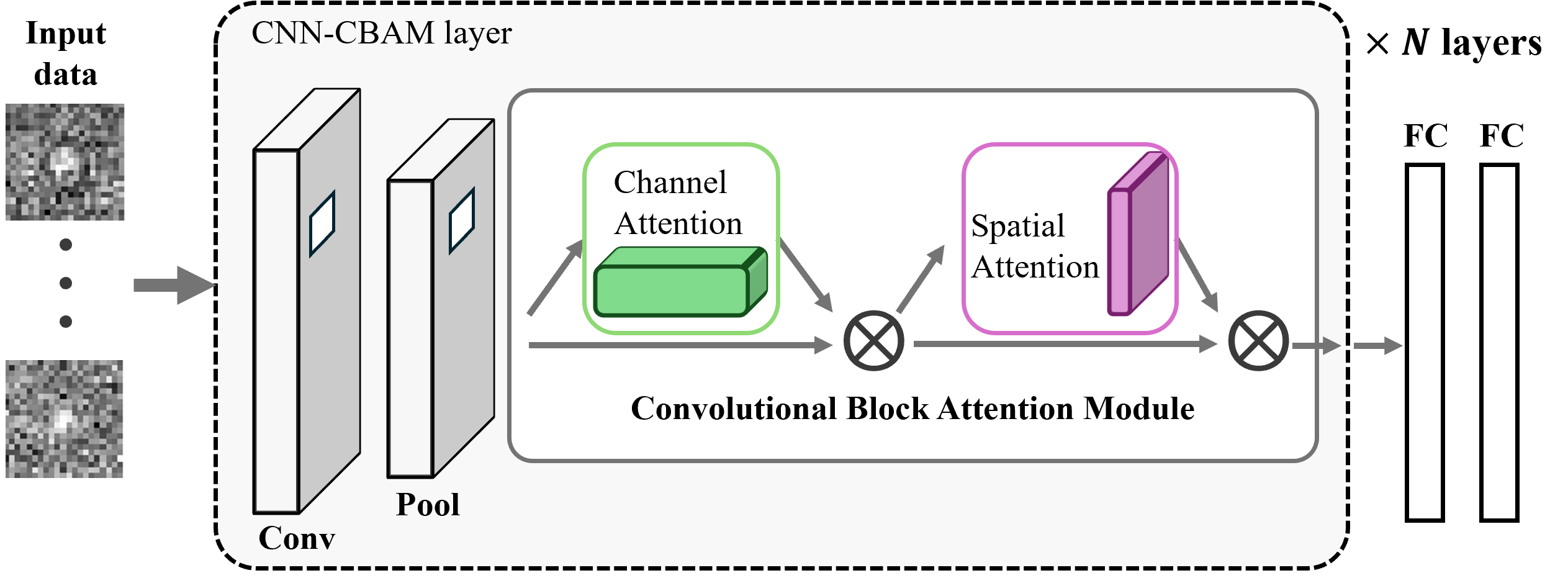}
  \caption{Architectures of the small-scale CNNs with CBAM. Each CNN-CBAM layer consists of one convolutional layer, one pooling layer, and one CBAM module. The models contain 2 or 4 such layers. {Alt text: Block diagrams of compact convolutional networks where each block includes convolution, pooling, and CBAM; variants stack two or four blocks before a classifier head producing a binary probability of object presence.}}
  \label{fig:CNN_CBAM}
\end{figure}

The small-scale CNN models vary in depth and channel configuration.
The smallest model uses two convolutional layers with channels 32 and 64, while the largest has four layers with channels 64, 128, 256, and 512.
The input size is $20 \times 20$ pixels. We apply $3 \times 3$ convolutions and $2 \times 2$ pooling to retain discriminative details and prevent the loss of fine features in small regions. 
Each model employs batch normalization, dropout, and ReLU activation.
The final layer uses a sigmoid function for binary classification. As shown in Figure~\ref{fig:CNN_CBAM}, each layer includes a CBAM module. Model configurations are summarized in Table~\ref{tab:cnn_models}.
In all small-scale CNN experiments, we trained all parameters from scratch. 
Specifically, all parameters, including those of the CBAM modules, were randomly initialized and optimized during training without any pretraining.

\begin{table}[htbp]
  \centering
  \caption{Detailed configurations of the small-scale CNN models.}
  \begin{tabular}{ccccc}
    \hline
    \multirow{2}{*}{Model ID} & \multicolumn{4}{c}{Number of Channels} \\
    & Layer 1 & Layer 2 & Layer 3 & Layer 4 \\
    \hline
    CNN1 & 32 & 64 & -- & -- \\
    CNN2 & 64 & 128 & -- & -- \\
    CNN3 & 32 & 32 & 64 & 64 \\
    CNN4 & 64 & 64 & 128 & 128 \\
    CNN5 & 32 & 64 & 128 & 256 \\
    CNN6 & 64 & 128 & 256 & 512 \\
    \hline
  \end{tabular}
  \label{tab:cnn_models}
\end{table}

\subsubsection{Large-scale CNNs}

\begin{figure}[htbp]
  \centering
  \includegraphics[width=\linewidth]{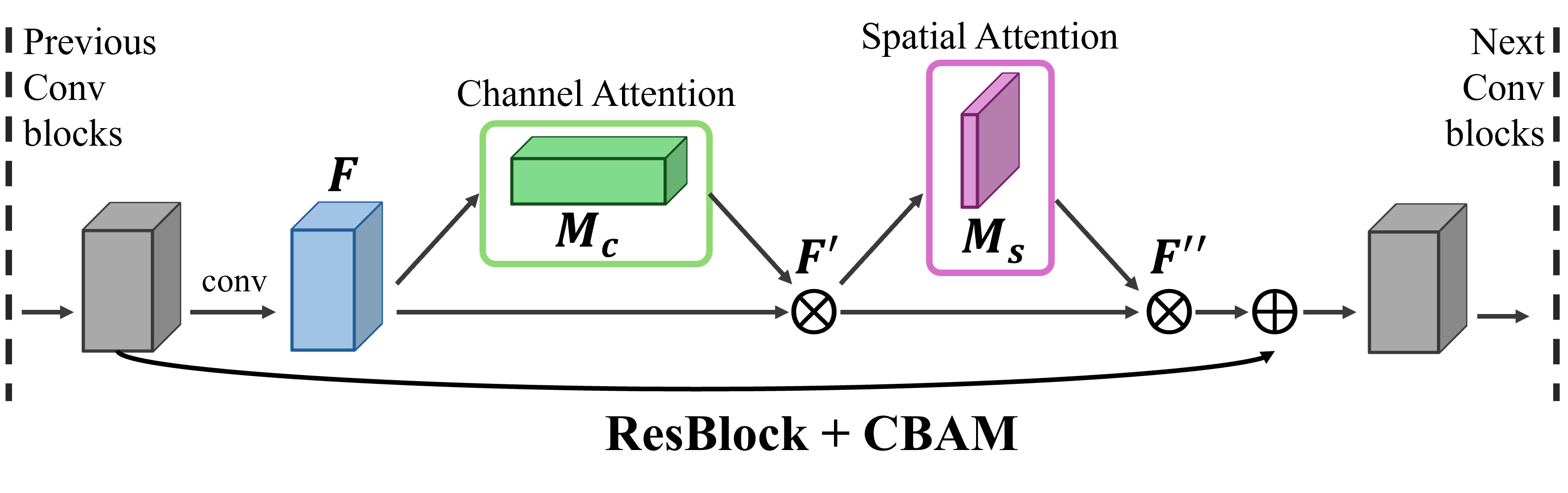}
  \caption{ResNet block architecture with CBAM. The attention module is inserted between residual blocks. {Alt text: Schematic of a residual block with a shortcut connection; CBAM is placed between consecutive residual blocks to compute channel and spatial weights that are multiplied with the intermediate feature maps.}}
  \label{fig:resnet_block}
\end{figure}

For large-scale models, we used ResNet18, ResNet34, ResNet50, ResNet101, and ResNet152. 
ResNet, proposed by Microsoft in 2015, is a deep CNN architecture that stabilizes training via skip connections, addressing the vanishing gradient problem \citep{he2015}. 
While CBAM has proven effective on color images, its impact on our multi-image stacked inputs remains uncertain, motivating a comparative study.
Given the smaller input size (\(32 \times 32\)), the original \(7 \times 7\) input filter was replaced with a \(3 \times 3\) kernel. 
Other architecture settings follow the original ResNet. In our implementation, CBAM is inserted between residual blocks. 
For example, ResNet50 consists of 16 three-layer residual blocks, each followed by a CBAM module. 
Figure~\ref{fig:resnet_block} illustrates a ResNet block with CBAM.
For all ResNet experiments, we first initialized the models with weights pretrained on a large-scale dataset ImageNet\citep{Deng2009}, and then fine-tuned all parameters using our training data. 
We did not freeze any layers; instead, all weights, including those of the CBAM modules, were updated during training.

\subsubsection{Adjustments for Overfitting}
Table~\ref{tab:data_param_ratio} lists the ratio of total samples to model parameters. 
CNN1 and CNN3 have relatively high ratios ($0.201$ and $0.101$), suggesting a lower risk of overfitting. 
CNN2, CNN4-6, and all ResNet models show markedly lower ratios, with ResNet152 reaching $0.000272$, indicating a high risk.
These criteria are usually based on three-channel Red-Green-Blue(RGB) inputs, whereas our dataset uses multi-channel images (5-32 channels). 
Because each sample contains more information than an RGB image, the actual effect on overfitting is uncertain.

We applied Batch Normalization\citep{ioffe2015} to every layer and pre-trained ResNet models on ImageNet\citep{Deng2009}. 
Due to the difference between ImageNet’s three-channel images and our data, all parameters were fine-tuned. 
We also introduced early stopping.

Early stopping is a regularization technique commonly used in training machine learning models to reduce overfitting. In this method, the dataset is split into training and validation subsets, and the model’s performance on the validation set is monitored throughout training. The parameters that yield the best validation performance are recorded, and training is halted when the validation metric stops improving for a predefined number of iterations. The model parameters are then reverted to those corresponding to the optimal validation performance. This approach helps prevent overfitting by stopping the learning process at the point where the model generalizes best.

\begin{table}[htbp]
\centering
\caption{Ratio of number of total samples to model parameters.}
\label{tab:data_param_ratio}
\begin{tabular}{lccc}
\hline
Model & Parameters & Total samples & Samples / Parameter \\ \hline
CNN1   & $4.7 \times 10^{4}$  & $9.4 \times 10^{3}$ & $2.0 \times 10^{-1}$ \\
CNN2   & $1.3 \times 10^{5}$  & $9.4 \times 10^{3}$ & $7.3 \times 10^{-2}$ \\
CNN3   & $9.3 \times 10^{4}$  & $9.4 \times 10^{3}$ & $1.0 \times 10^{-1}$ \\
CNN4   & $3.2 \times 10^{5}$  & $9.4 \times 10^{3}$ & $3.0 \times 10^{-2}$ \\
CNN5   & $5.0 \times 10^{5}$  & $9.4 \times 10^{3}$ & $1.9 \times 10^{-2}$ \\
CNN6   & $1.8 \times 10^{6}$  & $9.4 \times 10^{3}$ & $5.3 \times 10^{-3}$ \\ \hline
ResNet18  & $3.2 \times 10^{6}$  & $9.4 \times 10^{3}$ & $3.0 \times 10^{-3}$ \\
ResNet34  & $1.1 \times 10^{7}$  & $9.4 \times 10^{3}$ & $8.3 \times 10^{-4}$ \\
ResNet50  & $1.3 \times 10^{7}$  & $9.4 \times 10^{3}$ & $7.1 \times 10^{-4}$ \\
ResNet101 & $2.5 \times 10^{7}$  & $9.4 \times 10^{3}$ & $3.8 \times 10^{-4}$ \\
ResNet152 & $3.5 \times 10^{7}$  & $9.4 \times 10^{3}$ & $2.7 \times 10^{-4}$ \\ \hline
\end{tabular}
\end{table}

\subsection{CBAM: Convolutional Block Attention Module}\label{ssec:3.1}
\begin{figure}[htbp]
  \centering
  \includegraphics[width=\linewidth]{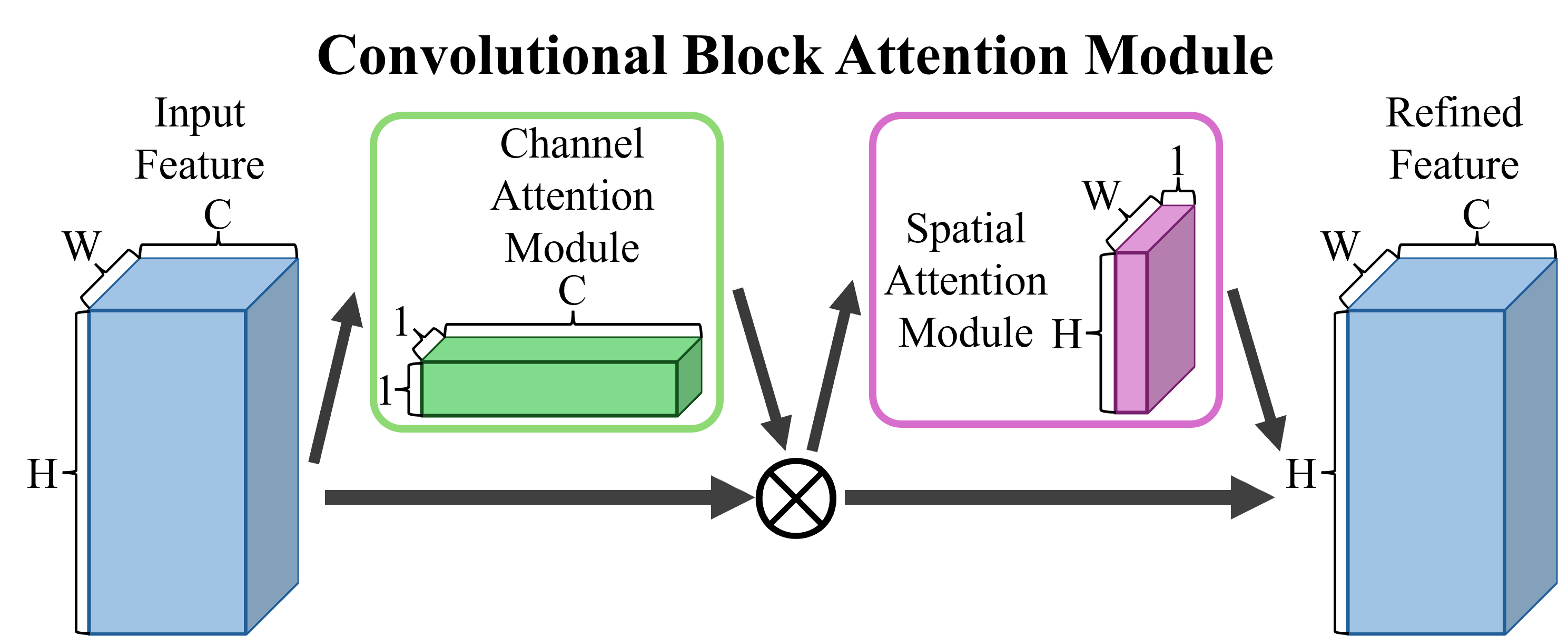}
  \caption{Convolutional Block Attention Module (CBAM) architecture. {Alt text: Diagram of CBAM showing sequential channel attention and spatial attention applied to a feature map; each attention produces a weight map that is multiplied element-wise with the features to refine salient information.}}
  \label{fig:cbam}
\end{figure}

The Convolutional Block Attention Module (CBAM) is an attention mechanism designed for Convolutional Neural Network \citep[CNN; ][]{woo2018}. 
CBAM can be integrated into existing CNN architectures to enhance feature representation and improve model performance.
By learning both channel-wise and spatial attention, CBAM highlights important features within the feature maps.
The channel attention module emphasizes significant channels, while the spatial attention module highlights relevant spatial regions. 
Figure~\ref{fig:cbam} illustrates the architecture of CBAM.
The input to CBAM is a three-dimensional feature map \( \mathbf{F} \in \mathbb{R}^{C \times H \times W} \), where \( C \), \( H \), and \( W \) denote the number of channels, height, and width, respectively. 
This feature map passes through both the channel and spatial attention modules, yielding an output with enhanced salient features. 
The process is formulated as:
\begin{equation}
  \begin{aligned}
  \mathbf{F}^{'}  &= M_c(\mathbf{F}) \otimes \mathbf{F}, \\
  \mathbf{F}^{''} &= M_s(\mathbf{F}^{'}) \otimes \mathbf{F}^{'},
  \end{aligned}
\end{equation}
where \(\otimes\) denotes element-wise multiplication, and \( M_c \) and \( M_s \) represent the channel and spatial attention modules, respectively.

\subsubsection{Channel Attention Module}
The channel attention module learns the significance of each channel in the feature map.
The top of Figure~\ref{fig:cam_sam} illustrates the architecture of the channel attention module.
It applies both average pooling and max pooling across spatial dimensions, then processes the results through a shared Multi-Layer Perceptron (MLP) consisting of one hidden layer with a Rectified Linear Unit(ReLU) activation.
The hidden layer has dimensionality \( \mathbb{R}^{C/r \times 1 \times 1} \), where \( r \) is the reduction ratio.
The two outputs are summed element-wise and passed through a sigmoid function to generate the final attention map. The process is defined as:
\begin{equation}
\begin{aligned}
M_c(\mathbf{F}) &= \sigma\left(\text{MLP}(\text{AvgPool}(\mathbf{F})) + \text{MLP}(\text{MaxPool}(\mathbf{F}))\right) \\
&= \sigma\left(W_1(W_0(\mathbf{F}_{\text{avg}}^c)) + W_1(W_0(\mathbf{F}_{\text{max}}^c))\right),
\end{aligned}
\end{equation}
where \( W_0 \in \mathbf{R}^{C/r \times C} \) and \( W_1 \in \mathbb{R}^{C \times C/r} \) are MLP weights, and \( \sigma \) is the sigmoid activation function.

\subsubsection{Spatial Attention Module}
\begin{figure}[htbp]
  \centering
  \includegraphics[width=\linewidth]{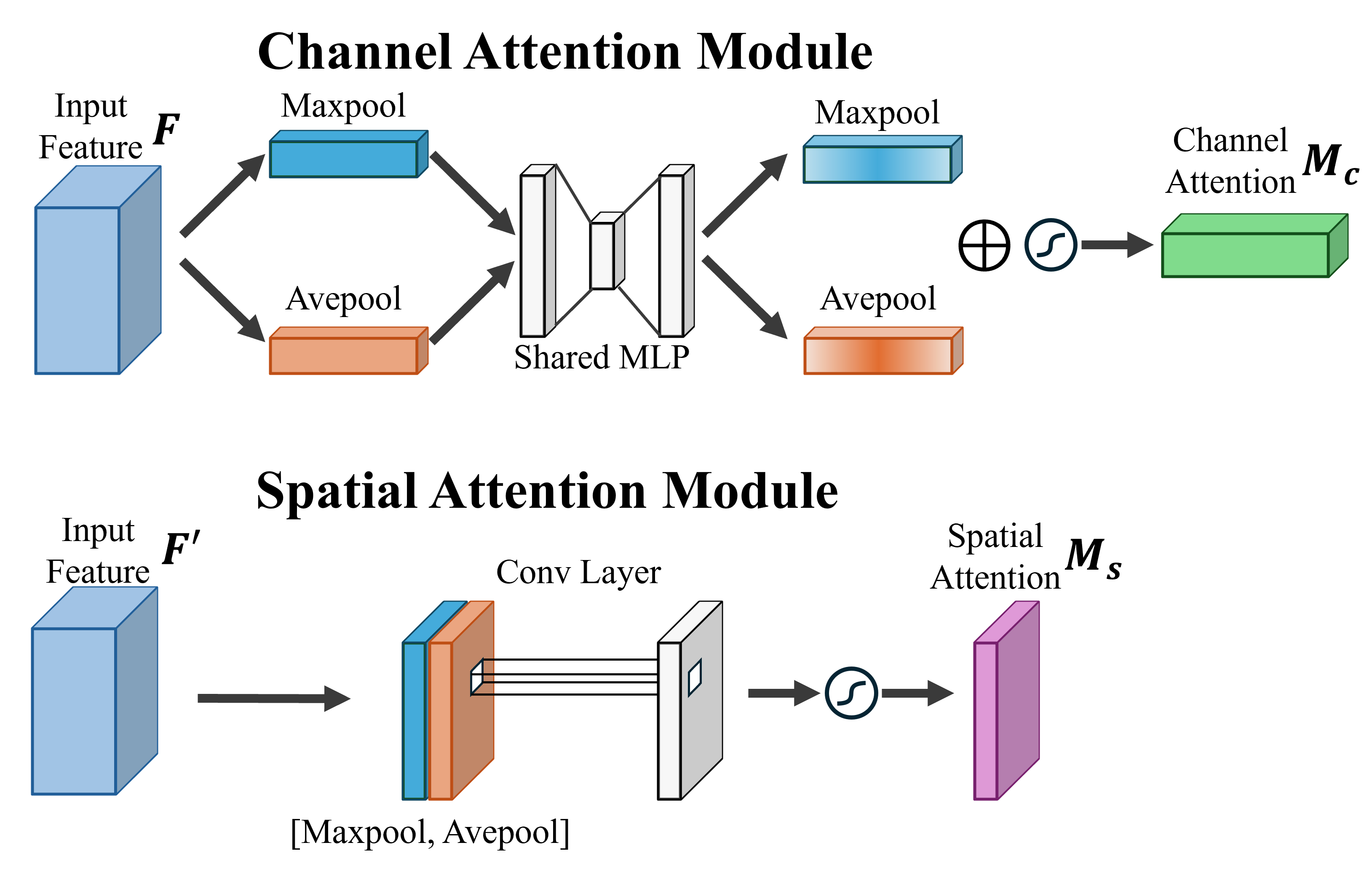}
  \caption{Architectures of the Channel Attention Module (top) and the Spatial Attention Module (bottom). {Alt text: Two diagrams. The channel attention diagram applies average and max pooling over spatial dimensions, feeds both through a shared multilayer perceptron, and produces channel weights. The spatial attention diagram pools over channels, concatenates the maps, and uses a seven-by-seven convolution to generate a spatial weight map.}}
  \label{fig:cam_sam}
\end{figure}

The spatial attention module determines which regions within the feature map are spatially significant.
The bottom of Figure~\ref{fig:cam_sam} illustrates the architecture of the spatial attention module.
It first performs average and max pooling along the channel axis, yielding two feature maps \( \mathbf{F}_{\text{avg}}^s \) and \( \mathbf{F}_{\text{max}}^s \), each of shape \( \mathbb{R}^{1 \times H \times W} \).
These are concatenated and passed through a \( 7 \times 7 \) convolutional layer to produce the spatial attention map:
\begin{equation}
\mathbf{M}_s(\mathbf{F}) = \sigma\left(f^{7 \times 7}([\mathbf{F}_{\text{avg}}^s; \mathbf{F}_{\text{max}}^s])\right),
\end{equation}
where \( f^{7 \times 7} \) denotes the convolution operation, and \( [\cdot ; \cdot] \) denotes concatenation along the channel axis.
CBAM refines intermediate feature maps by emphasizing meaningful information in both channel and spatial dimensions.
When integrated into CNNs, CBAM modules are typically inserted between convolutional blocks.
In the case of ResNet, CBAM is applied between residual blocks for optimal refinement.

\subsection{Metrics for Evaluation}\label{sssec:3.3} 
To assess the performance of the models, we employ five key evaluation metrics: accuracy, recall, precision, F1-score, and inverse precision.
Accuracy measures the proportion of correctly classified images and is particularly effective when the dataset maintains a balanced distribution of positive and negative samples.
Recall quantifies the proportion of actual objects that are correctly detected. 
This metric is crucial for minimizing missed detections thereby ensuring a comprehensive search range. 
Precision represents the proportion of detected objects that are actual objects, reflecting the reliability of the model’s positive predictions.
The complement of precision, defined as \( 1 - \text{precision} \), indicates the detection error rate for classified objects. 
The F1-score balances precision and recall, making it a suitable metric for object detection tasks where both false positives and false negatives must be minimized.
Although not commonly used, inverse precision is introduced in this study to evaluate the model’s reliability in correctly identifying non-object images. 
It measures the proportion of non-detected objects that are truly absent. 
The complement, given by \( 1 - \text{inverse precision} \), quantifies the detection error rate for classified non-objects. 
These metrics are calculated as follows:
\begin{equation}
  \begin{aligned}
  \text{Accuracy} & = \frac{TP+TN}{TP+FP+FN+TN} \\
  \text{Recall} & = \frac{TP}{TP+FN} \\
  \text{Precision} & = \frac{TP}{TP+FP} \\
  \text{F1-score} & = 2 \times \frac{\mathrm{Precision} \times \mathrm{Recall}}{\mathrm{Precision}+\mathrm{Recall}} \\
  \text{Inverse Precision} & = \frac{TN}{TN+FP} \\
  \end{aligned}
\end{equation}
where TP (True Positive) refers to the number of correctly classified objects, FP (False Positive) refers to the number of non-object regions incorrectly classified as objects, FN (False Negative) refers to the number of objects that were not detected, and TN (True Negative) refers to the number of non-object regions correctly classified. 
These metrics are derived from the confusion matrix, which consists of four fundamental components: true positives (TP), false positives (FP), false negatives (FN), and true negatives (TN).
\begin{figure}[htbp]
  \centering
  \includegraphics[width=\linewidth]{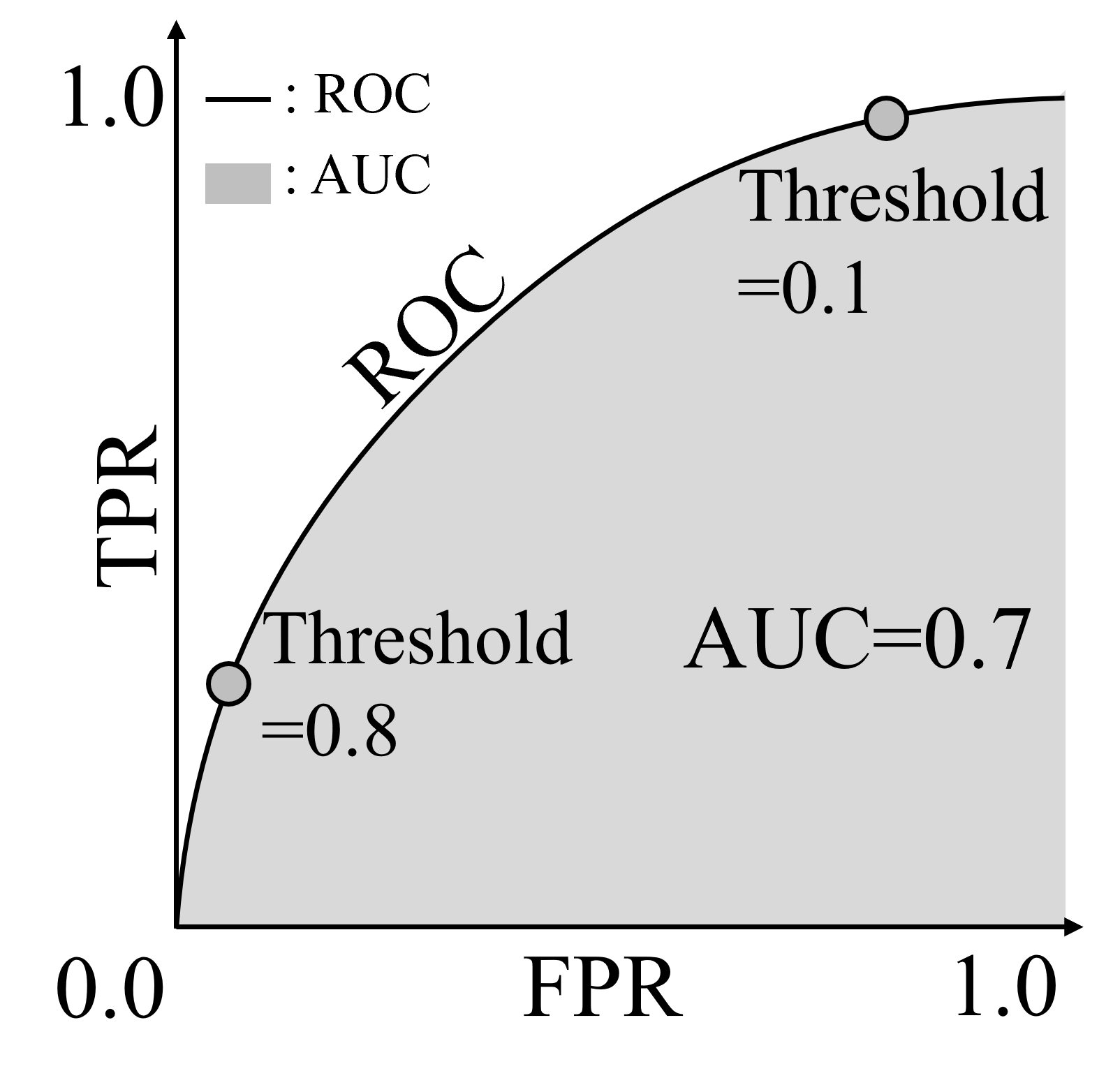}
  \caption{The concept of the AUC, a common metric for evaluating the performance of binary classification models. The ROC curve plots the True Positive Rate (TPR) against the False Positive Rate (FPR) at various threshold settings. {Alt text: Receiver operating characteristic plot with true positive rate versus false positive rate and example threshold points; the area under the curve summarizes class separability, where a larger area indicates better discrimination.}}
  \label{fig:auc}
\end{figure}

AUC (Area Under the Curve) is a threshold-independent metric that evaluates the overall ability of a classifier to distinguish between classes. 
It is especially useful for imbalanced datasets and situations where selecting a single decision threshold is challenging.
To compute AUC, we first consider the ROC (Receiver Operating Characteristic) curve, which plots the True Positive Rate (TPR) against the False Positive Rate (FPR) as the classification threshold varies from 0 to 1.
Each point on the ROC curve corresponds to a specific threshold. For example, Figure~\ref{fig:auc} shows ROC points at thresholds of 0.8 and 0.1.
The TPR and FPR are defined as:
\begin{equation}
\text{TPR} = \frac{\text{TP}}{\text{TP} + \text{FN}}, \quad
\text{FPR} = \frac{\text{FP}}{\text{FP} + \text{TN}}
\end{equation}
This area is typically computed using numerical integration.
\begin{itemize}
    \item \textbf{AUC = 1.0}: Represents perfect classification.
    This means the model can perfectly distinguish between positive and negative instances.
    The ROC curve would pass from (0,0) to (0,1) and then to (1,1).
    \item \textbf{AUC = 0.5}: Indicates that the classifier's performance is no better than random guessing.
    In this case, the ROC curve approximates the diagonal line from (0,0) to (1,1) (the line of no-discrimination or chance line).
    Such a classifier has no predictive value.
    \item \textbf{AUC < 0.5}: Suggests that the classifier's predictions are systematically incorrect, meaning misclassifications are dominant.
    In practice, inverting the classifier's predictions (i.e., swapping positive and negative labels) might improve performance.
\end{itemize}

\section{Experiment and Results\label{sec:result}}
In this section, we conduct two experiments to compare the classification performance of the 11 models described in Section \ref{sec:method}.
We also calculate the reduction rate of human costs when using the best-performing model and verify the effectiveness of ML methods.

\subsection{Experimental Settings}
All experiments follow the same training and evaluation protocol.
Performance is assessed via five‐fold cross‐validation, and we report the mean~$\pm$~standard deviation of each metric across the five folds.
All models are trained with the hyperparameters listed in Table~\ref{tab:exp_setup}, determined through a grid search to maximize the AUC.
We determined these hyperparameters through a grid search to maximize the AUC.

\begin{table}[htbp]
\centering
\caption{Experimental setup for CNN and ResNet model training}
\label{tab:exp_setup}
\resizebox{\linewidth}{!}{
\begin{tabular}{ll}
\hline
\textbf{Item} & \textbf{Description} \\ \hline
Threshold & 0.5 for all metrics except AUC \\
Optimizer & Adam (momentum = 0.9, weight decay = 0.0001) \\
Learning rate schedule & Initial LR = 0.001, reduced by factor 0.1 every 5 epochs \\
Epochs & 20 \\
Batch size & 32 \\
Loss function & Binary cross-entropy \\
\hline
\end{tabular}
}
\end{table}

\subsection{Experiment 1: Models with CBAM vs.\ without CBAM}
This preliminary experiment evaluates the effect of the Convolutional Block Attention Module (CBAM) by comparing each backbone \emph{with} CBAM against its corresponding \emph{without}-CBAM version.
The evaluation is restricted to uses only the data-combination scheme of 32 and 16 images.
Table~\ref{tab:cbam_compare} presents the results, contrasting the CBAM-enhanced variant with its baseline.
CBAM consistently improves both Accuracy and AUC for the tested backbones.
The AUC gains are notable (e.g., CNN2: $\Delta\text{AUC}=0.1827$), indicating enhanced ranking/separability.
The variance across folds remains controlled in CBAM models, suggesting that the improvements are robust to data splits.

\begin{table*}[htbp]
\centering
\caption{Comparison of CBAM and non-CBAM models using the 32–16 data combination in terms of Accuracy and AUC.}
\label{tab:cbam_compare}
\begin{tabular}{lcccc|cc}
\hline
Model & Accuracy (CBAM) & Accuracy (w/o) & AUC (CBAM) & AUC (w/o) & $\Delta$Accuracy & $\Delta$AUC \\ \hline
CNN1 & 0.9423±0.0198 & 0.9247±0.0369 & 0.9209±0.0453 & 0.8644±0.0675 & +0.0176 & +0.0565 \\
CNN2 & 0.9418±0.0315 & 0.8710±0.0439 & 0.9143±0.0632 & 0.7447±0.0890 & +0.0708 & +0.1696 \\
CNN3 & 0.9573±0.0168 & 0.9260±0.0369 & 0.9436±0.0315 & 0.9117±0.0688 & +0.0313 & +0.0319 \\
CNN4 & 0.9510±0.0200 & 0.9131±0.0516 & 0.9341±0.0387 & 0.9087±0.0582 & +0.0379 & +0.0254 \\
CNN5 & 0.9587±0.0163 & 0.9404±0.0275 & 0.9453±0.0294 & 0.9224±0.0488 & +0.0183 & +0.0229 \\
CNN6 & 0.9535±0.0197 & 0.9443±0.0217 & 0.9375±0.0424 & 0.9337±0.0375 & +0.0092 & +0.0038 \\
resnet18 & 0.9386±0.0165 & 0.8727±0.0678 & 0.9155±0.0331 & 0.7943±0.1633 & +0.0659 & +0.1212 \\
resnet34 & 0.9356±0.0168 & 0.8081±0.0704 & 0.9120±0.0334 & 0.6556±0.1907 & +0.1275 & +0.2564 \\
resnet50 & 0.9304±0.0173 & 0.8959±0.0382 & 0.9061±0.0352 & 0.8276±0.0833 & +0.0345 & +0.0785 \\
resnet101 & 0.9272±0.0265 & 0.8670±0.0371 & 0.8991±0.0454 & 0.8393±0.0741 & +0.0602 & +0.0598 \\
resnet152 & 0.9193±0.0415 & 0.8040±0.0446 & 0.8839±0.0761 & 0.7075±0.1700 & +0.1153 & +0.1764 \\
\hline
\end{tabular}
\end{table*}

\subsection{Experiment 2: Training Performance of Models with CBAM} \label{ssec:4.1}
\subsubsection{Concept and Configuration}
We evaluate six CNN models (CNN1--CNN6) and five ResNet models (ResNet18, ResNet34, ResNet50, ResNet101, ResNet152) using five data‐combination schemes (``32'', ``32 16'', ``32 16 8'', ``32 16 8 4'', and ``32 4'') of Dataset1.  
The metrics considered are accuracy, recall, precision, inverse precision, F1 score, and AUC.  
The error bars in Figures 8-12 represent the standard deviation among the folds, reflecting the variability of the model performance due to the data splits.

\subsubsection{Overall Classification Performance}
\begin{figure}
    \centering
    \includegraphics[width=\linewidth]{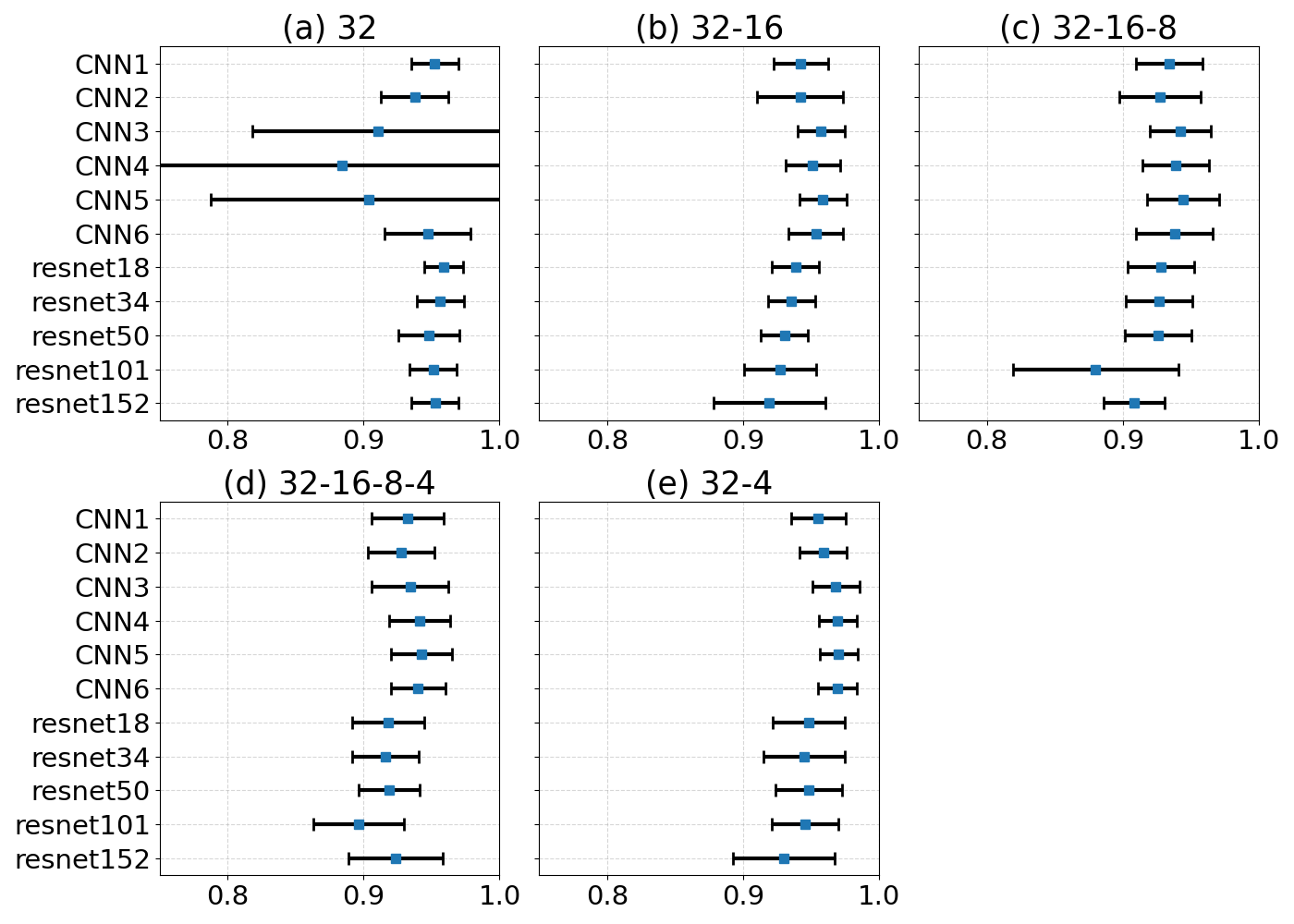}
    \caption{The Accuracy of each model across different input types, with error bars representing the standard deviation of the test-set Accuracy across the five folds of cross-validation. The y-axis indicates the models, and the x-axis shows the Accuracy values. Each subplot title specifies the combination of stacked images used as input. The blue dot within each error bar represents the mean Accuracy across the folds, and the bar length indicates the standard deviation among the folds. {Alt text: Matrix of error-bar plots showing accuracy by model and input combination; markers indicate fold means and horizontal bars show standard deviation across five folds for each configuration.}}
    \label{fig:exp1_acc}
\end{figure}

Accuracy is a highly intuitive metric for performance evaluation, particularly suitable for scenarios requiring general correctness of predictions.
Figure~\ref{fig:exp1_acc} displays the Accuracy for all models across data combinations.
For the ``32'' combination, ResNet18 achieved the highest accuracy ($0.9590\pm0.0158$), indicating its strong classification ability with single-resolution inputs.
For the multi-resolution combinations, CNN5 consistently exhibited superior accuracy, such as in ``32 16'' ($0.9587\pm0.0163$), ``32 16 8'' ($0.9446\pm0.0262$), and ``32 16 8 4'' ($0.9423\pm0.0215$).
Notably, in the ``32 4'' configuration, CNN5 reached the best performance ($0.9703\pm0.0149$), highlighting its robustness to varying input resolutions.
These results suggest that while ResNet18 is effective for simpler inputs, CNN5 excels as the complexity of input data increases.

\begin{figure}
    \centering
    \includegraphics[width=\linewidth]{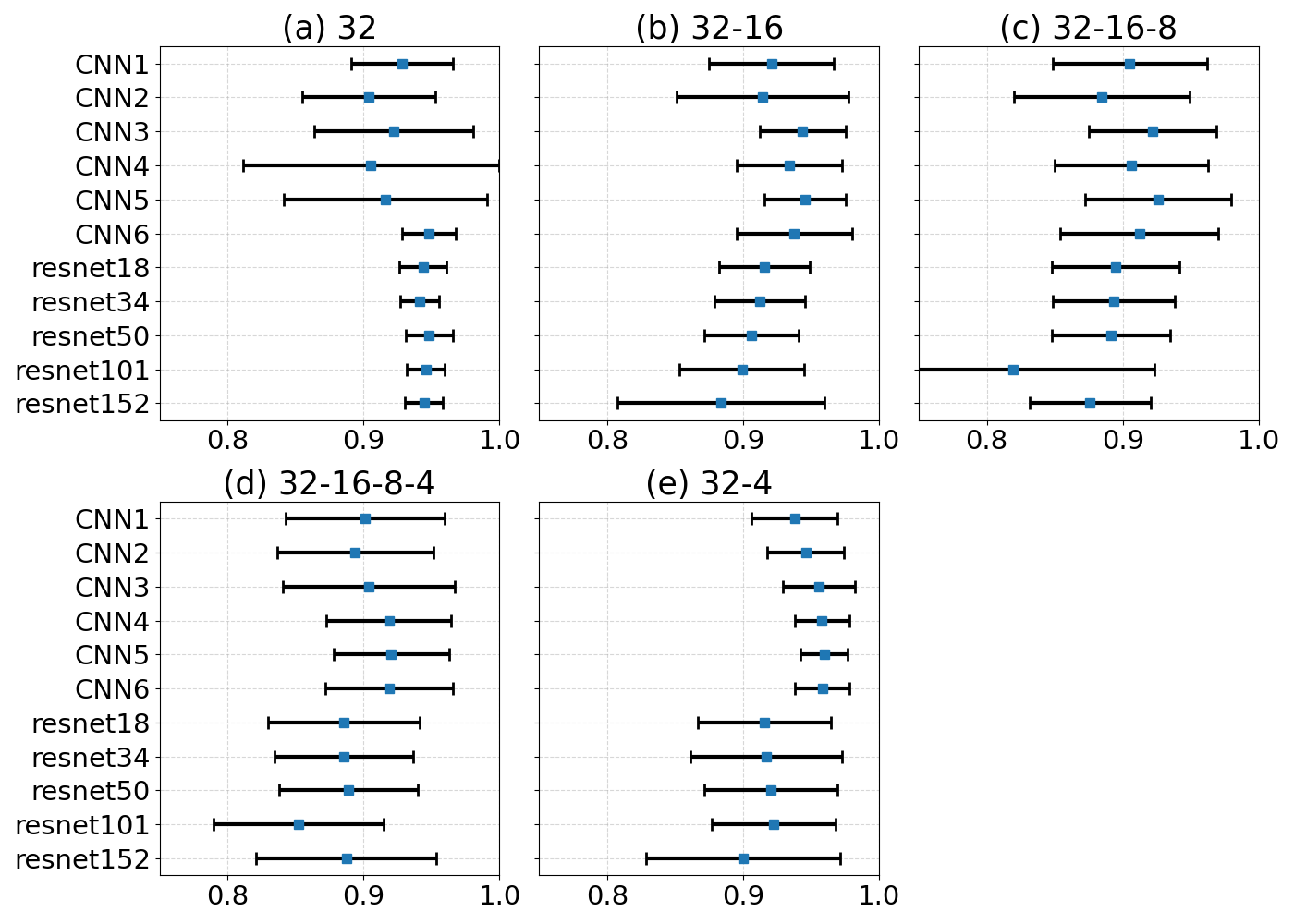}
    \caption{Evaluation of AUC for each model and data combination as demonstrated by the uncertainties. {Alt text: Matrix of error-bar plots of area under the ROC curve for multiple models and stacking configurations; higher values denote better ranking performance, with bars reflecting variability across cross-validation folds.}}
    \label{fig:exp1_auc}
\end{figure}

AUC is the primary metric for overall discriminative ability, being threshold-independent and robust to data imbalance.
Figure~\ref{fig:exp1_auc} displays the AUC for all models across data combinations.
For the ``32'' combination, ResNet models generally show high AUC, with ResNet50 achieving stable high performance (AUC $0.9483\pm0.0167$).
In contrast, for multi-resolution inputs (Figure~\ref{fig:exp1_auc}(b)--(d)), CNN models, particularly CNN5, consistently exhibit high AUC.
For instance, in ``32 16'' (Figure~\ref{fig:exp1_auc}(b)), CNN5 outperforms other models (AUC $0.9453\pm0.0294$), a trend also observed in ``32 16 8'' (AUC $0.9260\pm0.0536$) and ``32 16 8 4'' (AUC $0.9204\pm0.0426$).
For the ``32 4'' combination (Figure~\ref{fig:exp1_auc}(e)), CNN5 maintains high AUC ($0.9595\pm0.0162$), while CNN3 shows comparable and highly stable performance.
These results suggest that the optimal model architecture depends on the input data characteristics.
Visually, ResNet architectures tend to excel with single-resolution inputs, whereas specific CNNs (notably CNN5) perform better with multi-resolution inputs regarding AUC.
For the ``32 4'' combination has the highest AUC, indicating that the model can effectively distinguish between objects and non-objects across all input types.


\subsubsection{False Positive and Artifact Detection Analysis}
\begin{figure}
    \centering
    \includegraphics[width=\linewidth]{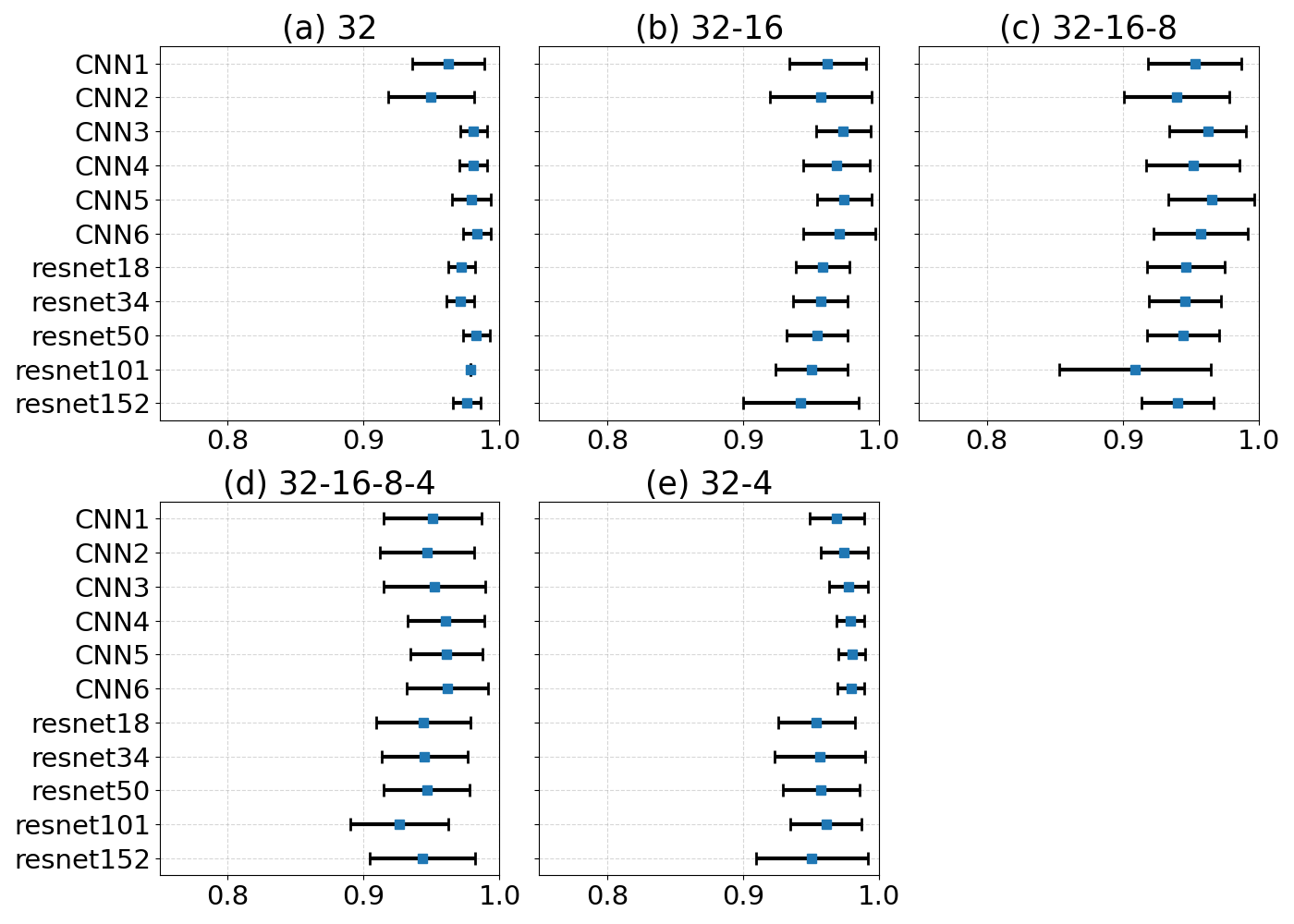}
    \caption{Evaluation of Precision for each model and data combination as demonstrated by the uncertainties. {Alt text: Error-bar charts of precision across models and input types; points show mean precision over folds and bars indicate variation, highlighting configurations that yield cleaner positive detections.}}
    \label{fig:exp1_pre}
\end{figure}

\begin{figure}
    \centering
    \includegraphics[width=\linewidth]{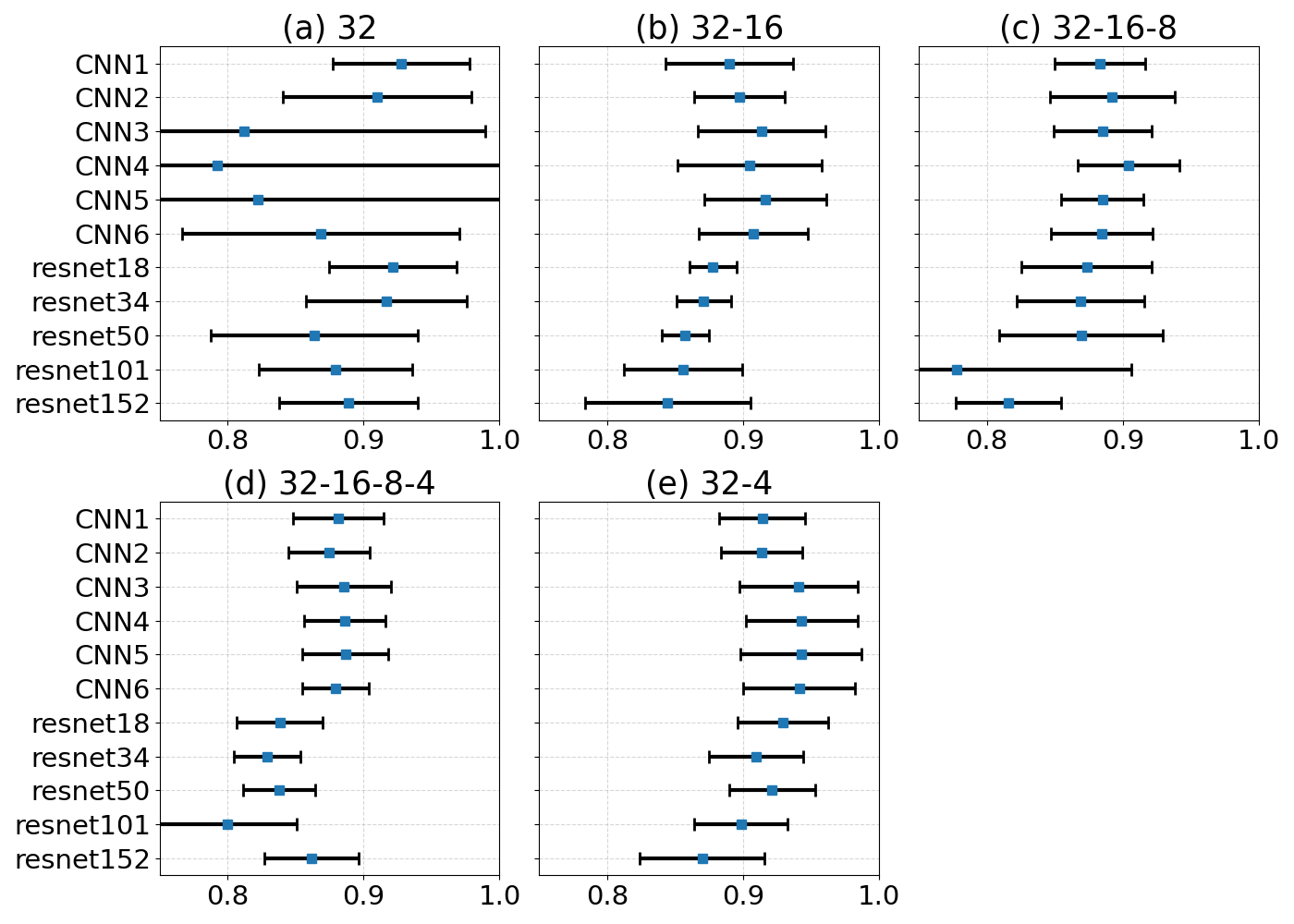}
    \caption{Evaluation of Inverse Precision for each model and data combination as demonstrated by the uncertainties. {Alt text: Error-bar charts of inverse precision, representing correctness on non-object images; larger values indicate fewer false positives among negatives, with bars showing cross-validation variability.}}
    \label{fig:exp1_invpre}
\end{figure}

We evaluate precision (accuracy for true objects among detected candidates) and inverse precision (accuracy for correctly identifying background/artifacts) to assess how cleanly models detect objects.
Figure~\ref{fig:exp1_pre} shows model precision, and Figure~\ref{fig:exp1_invpre} shows inverse precision.
For ``32'' (Figure~\ref{fig:exp1_pre}(a)), ResNet50 demonstrates the highest precision ($0.9829\pm0.0098$), indicating high confidence in its detected candidates.
With multi-resolution inputs, such as ``32 16'' (Figure~\ref{fig:exp1_pre}(b)) and ``32 4'' (Figure~\ref{fig:exp1_pre}(e)), CNN5 maintains high precision (e.g., $0.9799\pm0.0075$ for ``32 4'').

For inverse precision (Figure~\ref{fig:exp1_invpre}), which reflects the suppression of "junk" detections, CNN1 performs well for ``32'' (Figure~\ref{fig:exp1_invpre}(a), $0.9276\pm0.0495$), while CNN3 and CNN4 show large variability.
For multi-resolution inputs like ``32 16'' (Figure~\ref{fig:exp1_invpre}(b)), CNN5 achieves relatively high inverse precision ($0.9161\pm0.0449$).
``32 4'' (Figure~\ref{fig:exp1_invpre}(e)), CNN3 to CNN6, and especially CNN5 ($0.9426\pm0.0446$), show good inverse precision.

\subsubsection{Detection Yield}
\begin{figure}
    \centering
    \includegraphics[width=\linewidth]{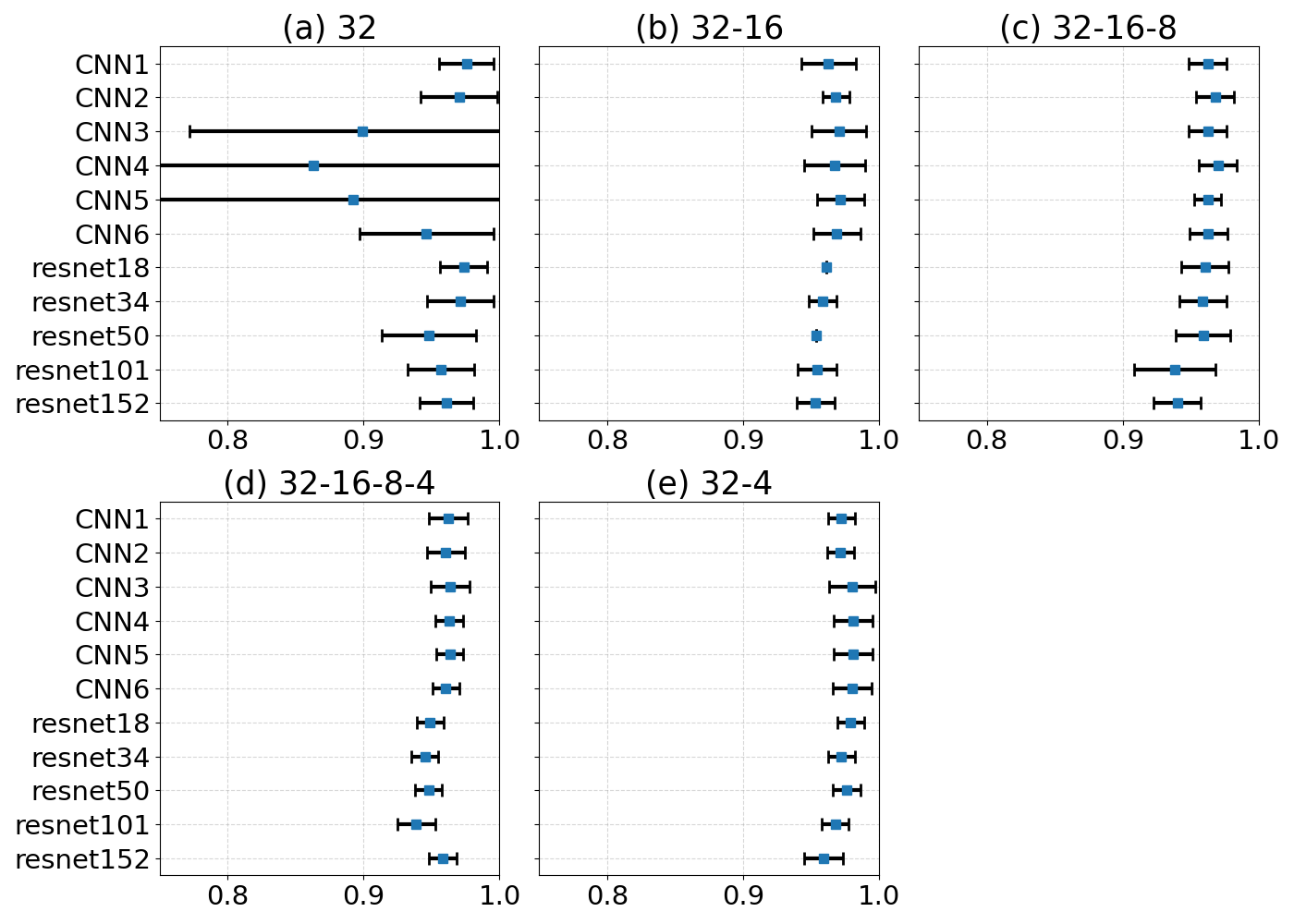}
    \caption{Evaluation of Recall for each model and data combination as demonstrated by the uncertainties. {Alt text: Error-bar charts of recall across models and inputs; higher values reflect greater detection yield of true objects, with variability shown by standard deviation bars across folds.}}
    \label{fig:exp1_rec}
\end{figure}

The practical efficiency of a survey is linked to its detection yield, i.e., how many true objects are detected, which is evaluated by recall.
Referring again to Figure~\ref{fig:exp1_rec}, models plotted further to the right (higher recall) exhibit higher detection efficiency.
For ``32'' (Figure~\ref{fig:exp1_rec}(a)), CNN1 and ResNet18 show high recall.
For multi-resolution inputs, CNN5 (``32 16''), CNN2/CNN4 (``32 16 8''), CNN1 (``32 16 8 4''), and CNN3/CNN5/CNN6 (``32 4'') all demonstrate high recall.

\subsubsection{Balanced Yield and Reliability}
\begin{figure}
    \centering
    \includegraphics[width=\linewidth]{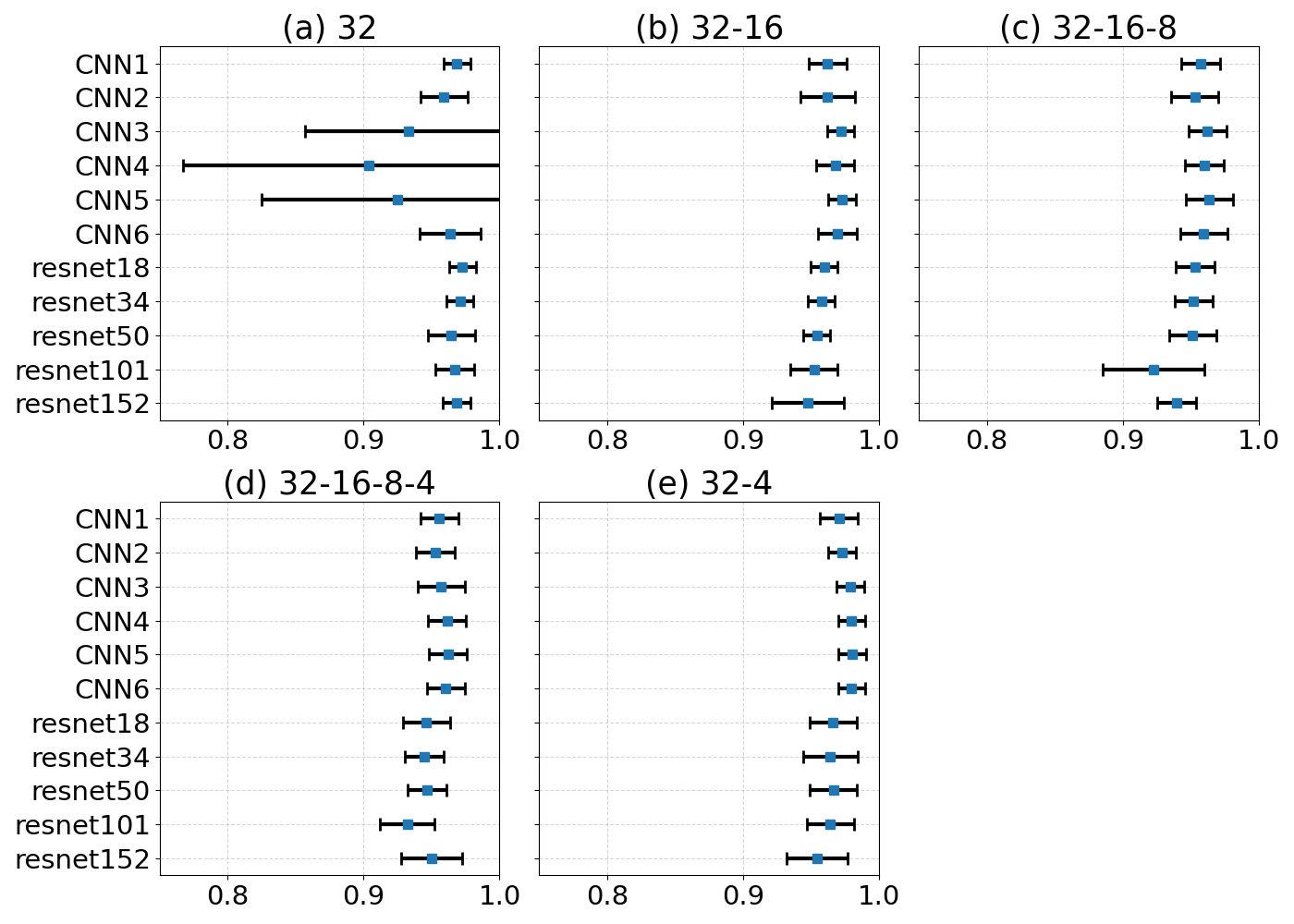}
    \caption{An error bar of F1 score for each model and data combination. {Alt text: Error-bar plots of F1 score summarizing the balance between precision and recall for each model–input configuration, with means and standard deviations across five cross-validation folds.}}
    \label{fig:exp1_f1}
\end{figure}

F1 is a balanced metric that combines both precision and recall, making it particularly valuable when false positives and false negatives must be simultaneously minimized.
Figure~\ref{fig:exp1_f1} summarizes the F1 scores across different data combinations.
In the ``32'' configuration, ResNet18 outperformed all other models with the highest F1 score ($0.9728\pm0.0107$), demonstrating strong and balanced detection performance.
CNN5 showed consistently high F1 scores across all multi-resolution settings, including ``32 16'' ($0.9726\pm0.0108$), ``32 16 8'' ($0.9635\pm0.0164$), and ``32 16 8 4'' ($0.9620\pm0.0136$).
The best F1 result overall was observed in the ``32 4'' setting with CNN5 ($0.9803\pm0.0100$), indicating exceptional effectiveness in diverse and complex input scenarios.
These trends affirm CNN5’s reliability in scenarios with intricate detection requirements.

\subsubsection{Summary of Data Shape and Model Selection}
The evaluation results demonstrate that model performance is decisively influenced by both the input data configuration and the network architecture.
While single-resolution inputs (e.g., ``32'') are best handled by deeper architectures such as ResNet18 and ResNet50 (i.e. achieving high AUC, F1, and precision), multi-resolution inputs allow lightweight CNNs, especially CNN5, to outperform deeper models across several evaluation metrics.

Among all tested configurations, the ``32 4'' input (i.e. combining the highest and lowest resolutions) achieved the best overall performance, with CNN5 recording the highest F1 score ($0.9803\pm0.0100$) and AUC ($0.9595\pm0.0162$).
This result highlights the benefit of integrating cross-scale spatial features, enabling even compact models to deliver robust and precise object detection.

Thus, multi-resolution inputs (especially ``32 4'') combined with CNN5 yield the most effective and stable performance in complex classification settings, whereas ResNet18 or ResNet50 remain optimal for simpler, single-resolution inputs.
Notably, model complexity alone does not guarantee performance.
In other words, what matters is the alignment between model architecture and the structure of the input data.

Considering overall classification performance (e.g., AUC in Figure~\ref{fig:exp1_auc} and accuracy in Figure~\ref{fig:exp1_acc}), the optimal model should be selected based on the scientific goal of the task:
\begin{itemize}
    \item \textbf{Initial Screening (maximize candidate detection):} Prioritize high recall (Figure~\ref{fig:exp1_rec}).
    Suitable choices include CNN1 (``32'', ``32 16 8 4'') and CNN5 (``32 16'').
    \item \textbf{High-Confidence Lists (minimize FPs/artifacts):} Prioritize high precision (Figure~\ref{fig:exp1_pre}) and inverse precision (Figure~\ref{fig:exp1_invpre}).
    Recommended models include ResNet50 (``32'') and CNN5 (``32 16'', ``32 4'').
    \item \textbf{Balanced Performance:} Prioritize high F1 score (Figure~\ref{fig:exp1_f1}).
    ResNet18 (``32'') and CNN5 (multi-resolution inputs) are top-performing candidates.
\end{itemize}

\subsection{Experiment 3: Evaluation Performance of Models with CBAM}\label{ssec:4.2}
\subsubsection{Concept and Configuration}
The classification performance of the trained model is evaluated on the unseen Dataset2.
Inference is carried out by ensembling the predictions of the $k$ models from the five‐fold cross‐validation in Experiment 2 using a majority‐voting scheme.
Evaluation metrics comprise accuracy, recall, precision, inverse precision, F1 score, and AUC.

\subsubsection{Overall Classification Performance}
\begin{figure}
    \centering
    \includegraphics[width=\linewidth]{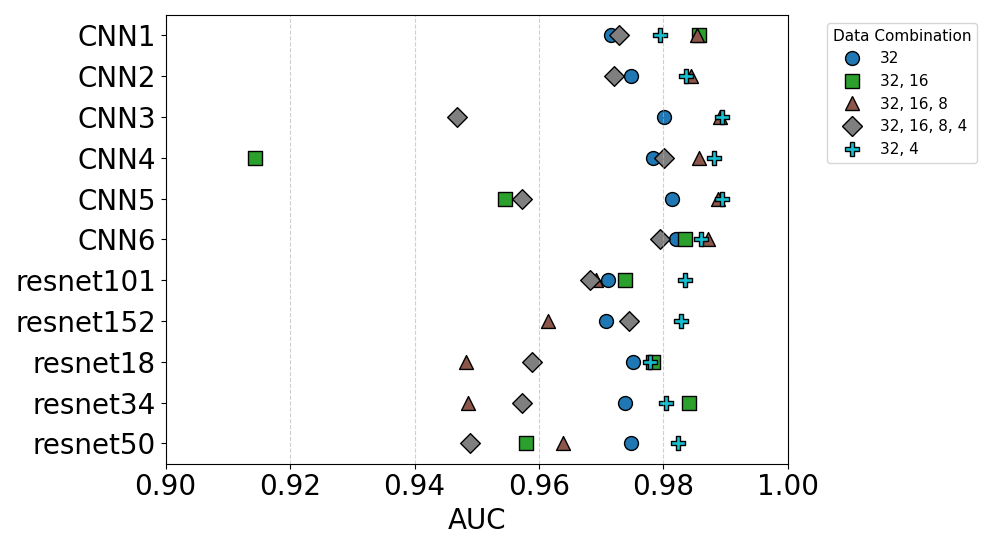}
    \caption{Scatter plot illustrating the relationship between each model-input combination and the corresponding Accuracy value. The y-axis represents the different models. Blue circles indicate input data using only the 32-frame stacked images. Green squares represent the combination of 32-frame and 16-frame stacked images. Red triangles correspond to the combination of 32-, 16-, and 8-frame stacked images. Gray diamonds indicate the use of 32-, 16-, 8-, and 4-frame stacked images. Blue crosses denote the combination of 32-frame and 4-frame stacked images. Some models do not include results for all data combinations because the value for those combinations fell below 0.90. {Alt text: Scatter plot of evaluation accuracy by model; marker shapes and colors encode different stacking combinations, enabling comparison of accuracy across architectures and input types.}}
    \label{fig:exp2_acc}
\end{figure}

Figure~\ref{fig:exp2_acc} displays the Accuracy for all models across data combinations from evaluation data.
For the ``32'' combination, CNN3 achieved the highest accuracy ($0.9808$), indicating strong classification capability for simpler input.
For multi-resolution inputs, performance remained high across various architectures.
Notably, ResNet34 outperformed other models in ``32, 16'' ($0.9851$), while CNN6 was the top performer in ``32, 16, 8'' ($0.9874$).
In ``32, 16, 8, 4'', CNN4 showed the highest accuracy ($0.9844$).
For the ``32, 4'' combination, CNN3 again achieved top performance with the highest overall accuracy ($0.9887$).
These results suggest that model choice should be tailored to input structure, with CNN3 and CNN6 demonstrating superior generalization in evaluation.

\begin{figure}
    \centering
    \includegraphics[width=\linewidth]{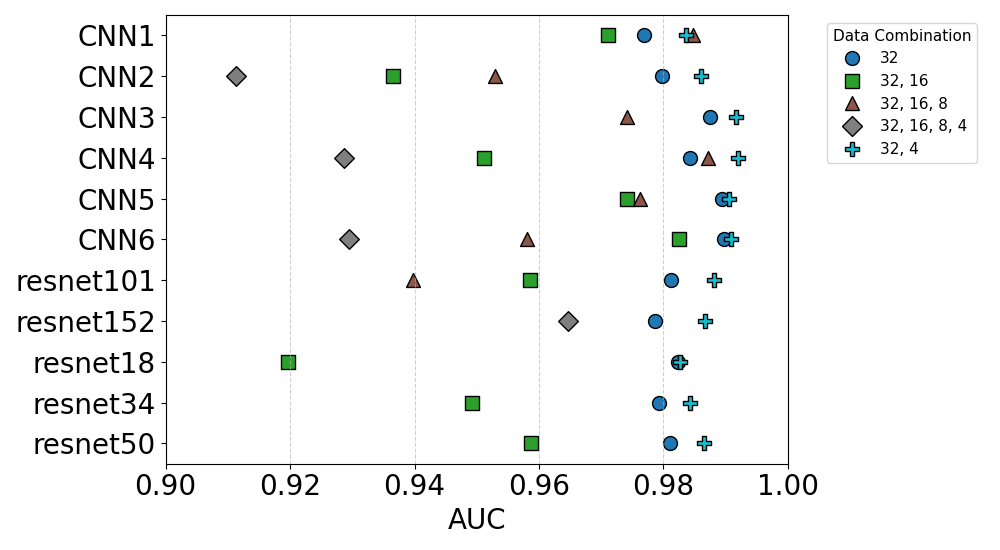}
    \caption{Scatter plot illustrating the relationship between each model-input combination and the corresponding AUC value. {Alt text: Scatter plot of AUC on the evaluation set for multiple models; distinct marker styles denote input combinations, summarizing overall discrimination performance across architectures.}}
    \label{fig:exp2_auc}
\end{figure}

Figure~\ref{fig:exp2_auc} displays the AUC for all models across data combinations from cross-validation.
For the ``32'' combination, ResNet models generally show high AUC, with ResNet50 achieving stable high performance ($0.9792$).
In contrast, for multi-resolution inputs, CNN models, particularly CNN5, consistently exhibit high AUC.
For the ``32 4'' combination, CNN5 maintains high AUC ($0.9901$), while CNN3 shows comparable and highly stable performance($0.9924$).
Again, the 32+4 combination has the highest AUC, indicating that the model can effectively distinguish between objects and non-objects across all input types.

\subsubsection{False Positive and Artifact Detection Analysis (Cross-Validation)}
\begin{figure}
    \centering
    \includegraphics[width=\linewidth]{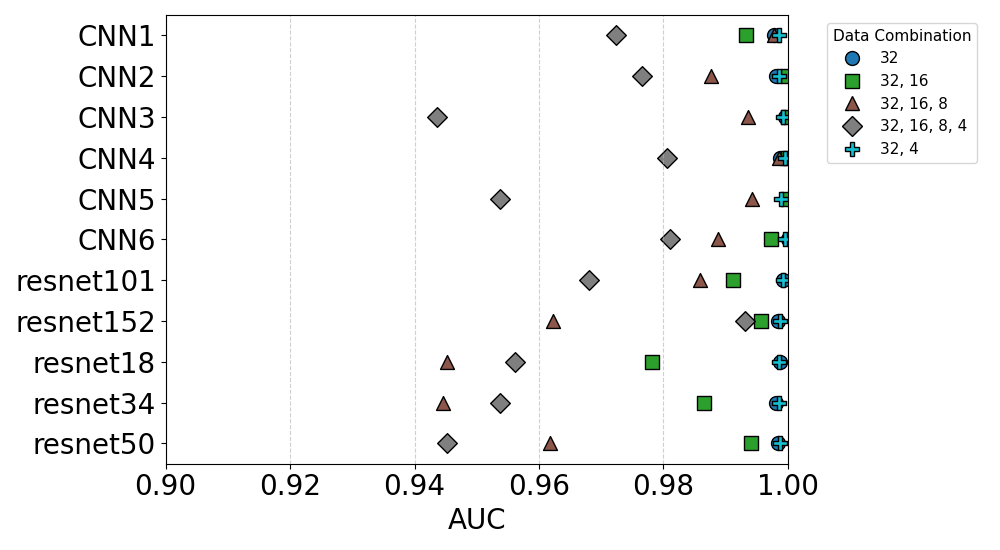}
    \caption{Scatter plot illustrating the relationship between each model-input combination and the corresponding Precision value. {Alt text: Scatter plot of precision across model–input combinations with markers indicating stacking types; higher points indicate cleaner positive candidate lists with fewer false positives.}}
    \label{fig:exp2_pre}
\end{figure}

\begin{figure}
    \centering
    \includegraphics[width=\linewidth]{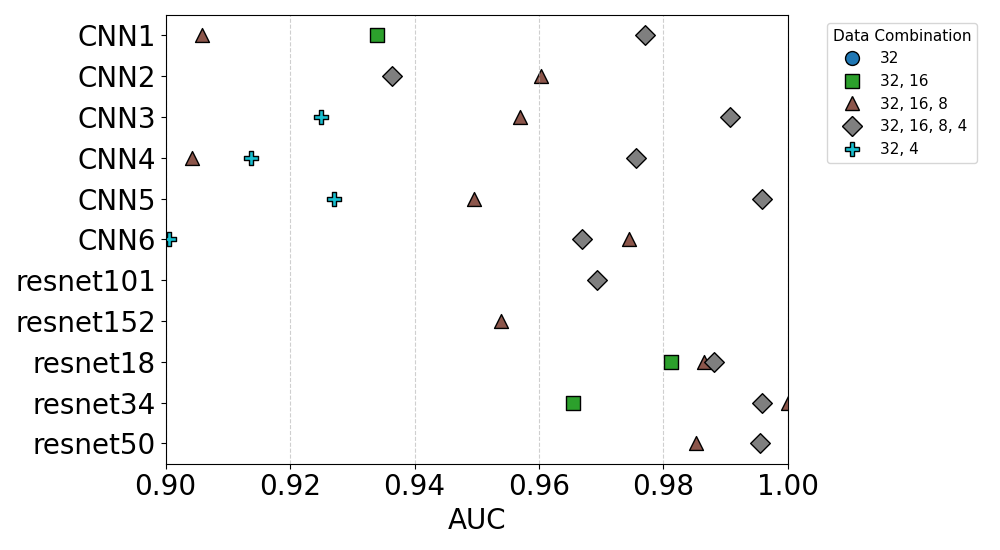}
    \caption{Scatter plot illustrating the relationship between each model-input combination and the corresponding Inverse Precision value. {Alt text: Scatter plot of inverse precision for evaluation data; marker encoding shows input combinations, and higher values indicate more reliable identification of non-object images.}}
    \label{fig:exp2_invpre}
\end{figure}

Figure~\ref{fig:exp2_pre} shows model precision, and Figure~\ref{fig:exp2_invpre} shows inverse precision from cross-validation.  
For ``32'', ResNet50 demonstrates the highest precision ($0.9996$).  
With multi-resolution inputs, such as ``32 4'', CNN5 maintains high precision ($0.9989$).  
Regarding inverse precision, CNN1 performs well for ``32 16 8 4'' ($1.000$).
However, the AUC is relatively low ($0.6801$), likely due to overfitting.
For ``32 4'', CNN3 to CNN6, especially CNN5 ($0.9225$), show good inverse precision.  
Models with high scores on both metrics are expected to yield reliable and clean detection lists.

\subsubsection{Detection Yield}
\begin{figure}
    \centering
    \includegraphics[width=\linewidth]{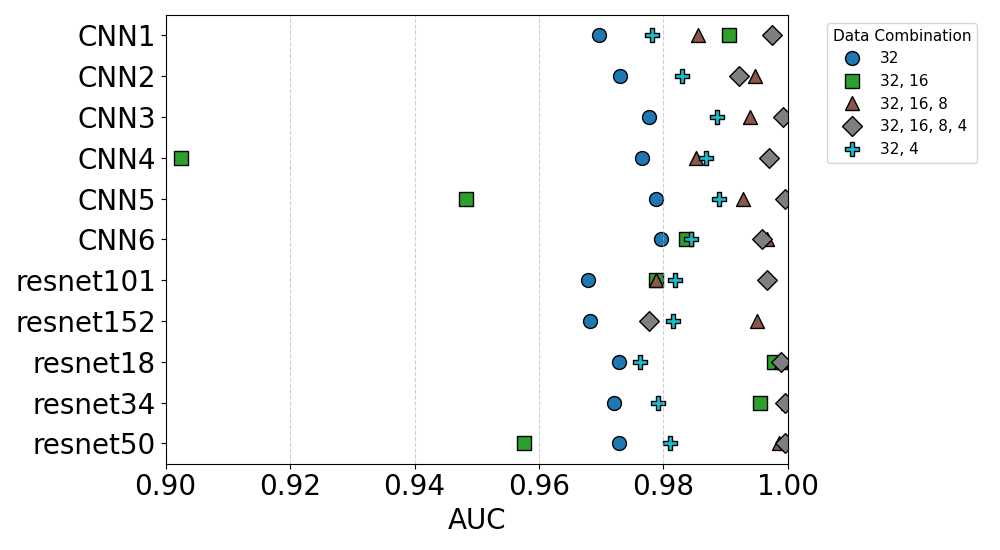}
    \caption{Scatter plot illustrating the relationship between each model-input combination and the corresponding Recall value. {Alt text: Scatter plot of recall across models and input combinations; higher values reflect better detection yield of true objects, facilitating comprehensive screening.}}
    \label{fig:exp2_rec}
\end{figure}

Referring to Figure~\ref{fig:exp2_rec} from cross-validation, models plotted further to the right (higher recall) exhibit higher detection efficiency.  
For ``32'' (Figure~\ref{fig:exp2_rec}(a)), CNN1 and ResNet18 show high recall.  
For multi-resolution inputs, CNN5 (``32 16''), and CNN3/CNN5/CNN6 (``32 4'') demonstrate high recall.

\subsubsection{Balanced Yield and Reliability}
\begin{figure}
    \centering
    \includegraphics[width=\linewidth]{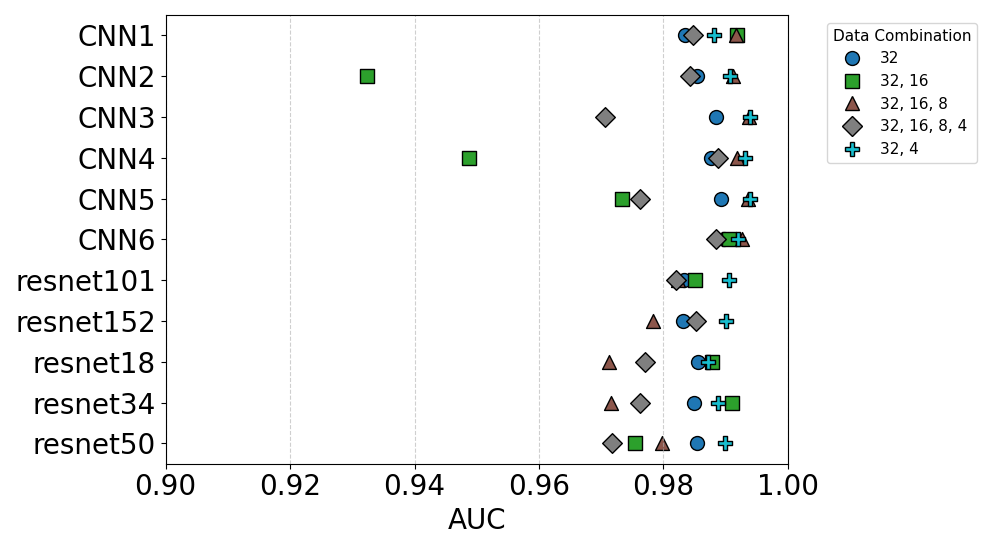}
    \caption{Scatter plot illustrating Recall for each model and data combination. {Alt text: Scatter plot of F1 score by model and input type on the evaluation set; marker styles represent stacking configurations, summarizing the balance between precision and recall.}}
    \label{fig:exp2_f1}
\end{figure}

Figure~\ref{fig:exp2_f1} displays the F1 scores for all models across evaluation data combinations.  
For the ``32'' setting, CNN3 exhibited the best F1 score ($0.9889$), highlighting its balance of precision and recall.  
Among multi-resolution inputs, ResNet34 led the ``32, 16'' combination ($0.9915$), while CNN6 performed best in ``32, 16, 8'' ($0.9928$).  
CNN4 achieved the highest score in the ``32, 16, 8, 4'' configuration ($0.9911$).  
Most notably, CNN3 recorded the best overall F1 score in the ``32, 4'' setting ($0.9935$), reflecting exceptional detection capability under more complex input arrangements.  
These results reinforce the suitability of CNN3 and CNN6 in robust classification tasks across varied resolutions.

\subsubsection{Summary of Data Shape and Model Selection}
The evaluation results confirm that model performance is closely tied to the input data structure.  
For single-resolution input (``32''), CNN3 achieves consistently high scores across accuracy ($0.9808$), AUC, and F1 ($0.9889$), indicating strong generalization with simpler input.  
However, multi-resolution strategies consistently improve performance and robustness across models.
In particular, the ``32, 4'' combination is highly successful.

CNN3 and CNN5 stand out for their stability and performance across multiple metrics.  
CNN3 achieved the highest overall accuracy ($0.9887$) and F1 score ($0.9935$) in ``32, 4'', while CNN5 maintained high AUC ($0.9595\pm0.0162$) and inverse precision ($0.9426\pm0.0446$).  
ResNet models, such as ResNet34, showed strength in ``32, 16'' settings, with notable F1 ($0.9915$), but CNN models often surpassed them in more diverse input scenarios.

\subsubsection{Overfitting Assessment}
\begin{figure}
    \centering
    \includegraphics[width=\linewidth]{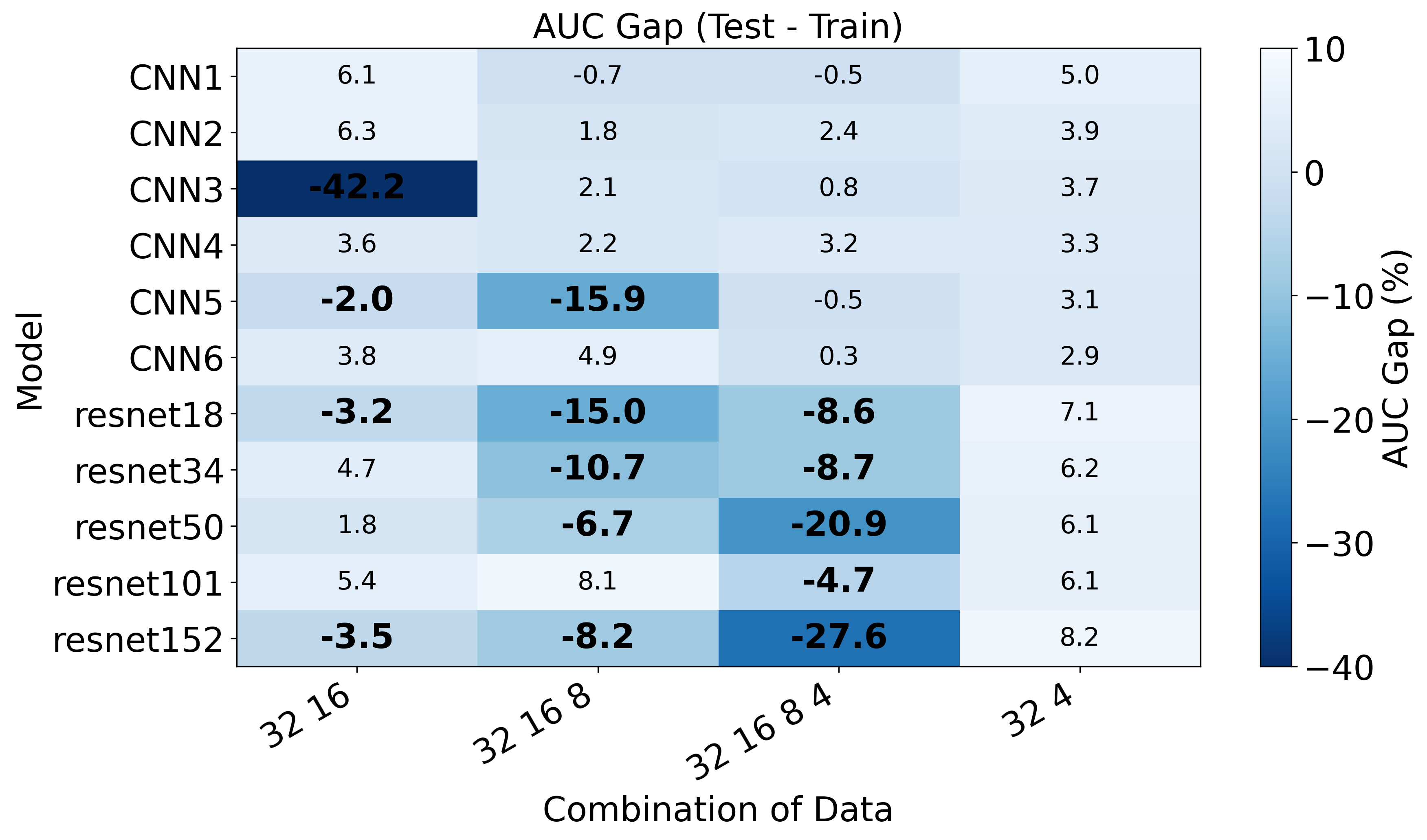}
    \caption{AUC generalization gap ($\mathrm{AUC}_{\text{test}} - \mathrm{AUC}_{\text{train}}$) for each model–data combination. Negative values indicate higher training performance than test performance, suggesting overfitting {Alt text: Heatmap of the AUC generalization gap by model and data combination; negative regions indicate overfitting where test AUC is lower than training AUC, with color intensity representing gap magnitude.}}
    \label{fig:auc_gap_heatmap}
\end{figure}

We quantified overfitting by the AUC generalization gap
$\Delta \mathrm{AUC} = \mathrm{AUC}_{\text{test}} - \mathrm{AUC}_{\text{train}}$ 
(Fig.~\ref{fig:auc_gap_heatmap}). Negative values indicate that training performance exceeds test performance.
As we increase the number of combined inputs (from \texttt{32 16} to \texttt{32 16 8 4}), several models exhibit clear overfitting. 
The most severe case appears in \texttt{CNN3} with \texttt{32 16} ($\Delta \mathrm{AUC}=-42.2\%$). 
ResNet also overfit under richer input combinations: \texttt{ResNet18} ($-15.0\%$ at \texttt{32 16 8}), \texttt{ResNet34} ($-10.7\%$ at \texttt{32 16 8}), \texttt{ResNet50} ($-20.9\%$ at \texttt{32 16 8 4}), and \texttt{ResNet152} ($-27.6\%$ at \texttt{32 16 8 4}). 
Among the plain CNNs, \texttt{CNN5} also shows substantial overfitting ($-15.9\%$ at \texttt{32 16 8}).
In contrast, many configurations yield gaps near zero (e.g., \texttt{CNN1}, \texttt{CNN2}, and \texttt{CNN6} around $-1\%$ to $+6\%$), which suggests adequate generalization.
These patterns align with the sample–to–parameter ratios in Table~\ref{tab:data_param_ratio}: models with larger capacity (especially deeper ResNets) tend to overfit when we increase input dimensionality via multi–combination stacks.
Our countermeasures reduce but do not eliminate overfitting in the largest–combination settings, likely due to the low samples–per–parameter regime and domain mismatch between RGB pretraining and multi–channel inputs.

\subsubsection{Calculation of Task Reduction Ratio}
In order to minimize the model’s classification error and at the same time reduce the human task, it is necessary to set appropriate thresholds that define the areas of responsibility between the model and the human.  
Two thresholds (positive and negative) are set to classify images.  
If the positive threshold is set and the output exceeds that value, the image is predicted to be an object.  
Similarly, if the negative threshold is set and the output is below that value, the image is classified as a false positive.  
In both cases, no human intervention is required. Only images with an output between the thresholds values are manually inspected.  
The thresholds are determined based on a trade-off between model error and task reduction rate.

\begin{figure}[htbp]
  \centering
  \includegraphics[width=\linewidth]{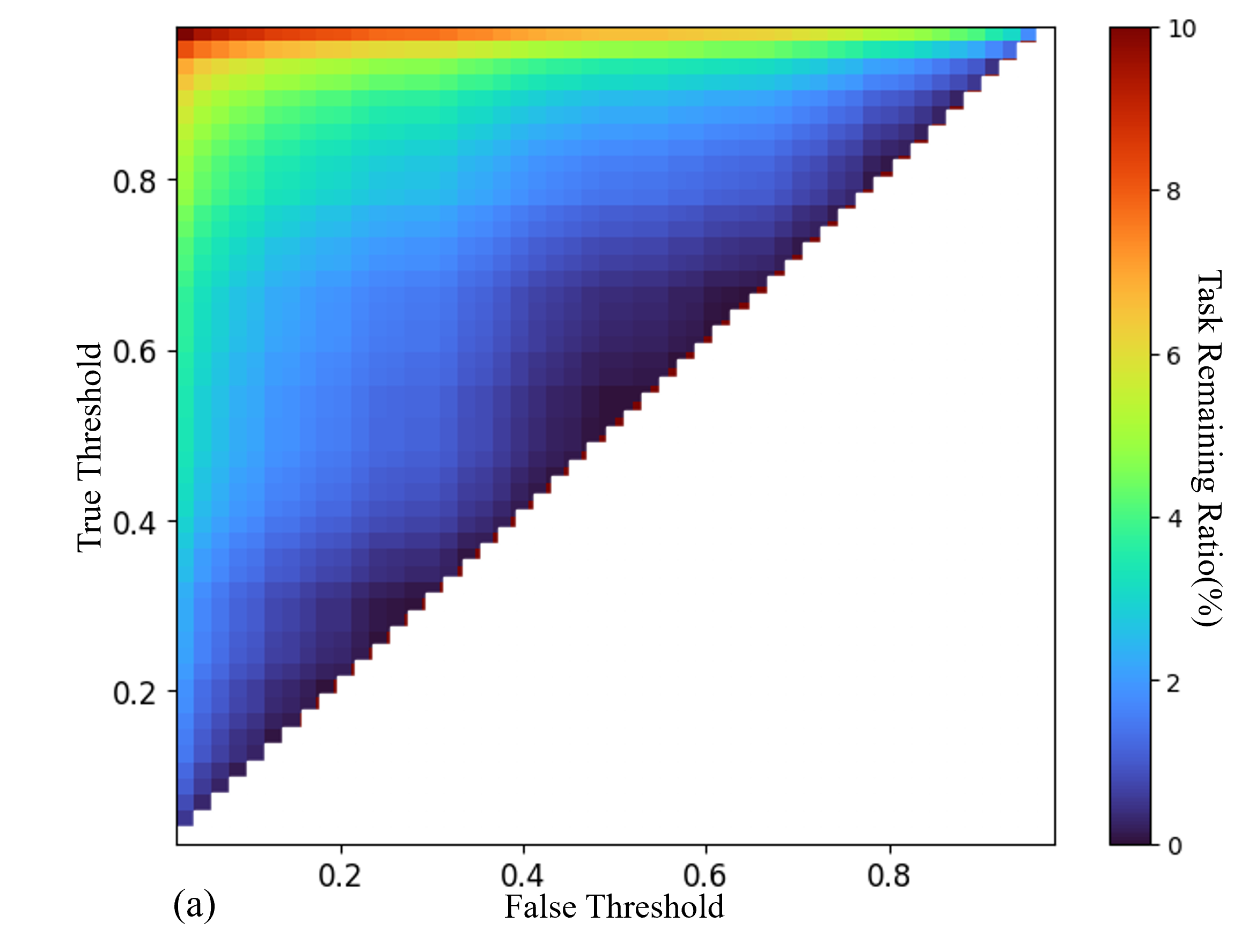}
  \includegraphics[width=\linewidth]{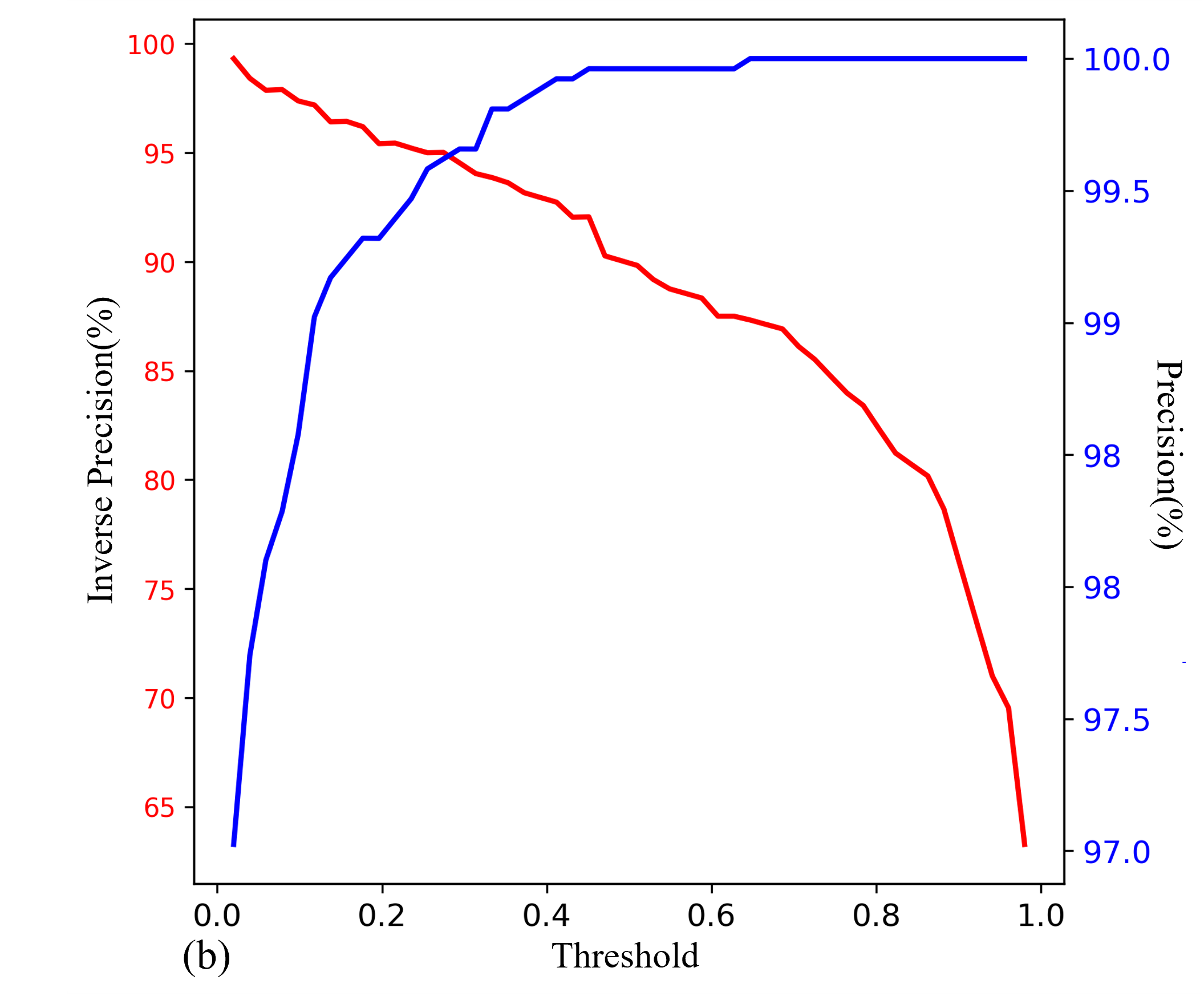}
  \caption{(a) Remaining task ratio (\%) across the positive and negative threshold combinations. Lower values (blue regions) indicate fewer remaining tasks, i.e., higher reduction of manual inspection. The white area represents infeasible threshold combinations where the true threshold is lower than the false threshold. (b) Graph of precision vs. positive threshold (blue line) and inverse precision vs. negative threshold (red line). {Alt text: Two-panel figure. The first panel shows a heatmap of remaining task ratio over pairs of positive and negative thresholds with an infeasible region blanked. The second panel shows curves of precision versus positive threshold and inverse precision versus negative threshold used to choose operating points.}}
  \label{fig:tradeoff}
\end{figure}

For the prediction results of the CNN3 model using the combination data for ``32, 4'' in Experiment 3, we comprehensively set positive and negative thresholds and perform a grid search for precision, inverse precision, and task reduction rate at each threshold.  
The results are shown in Fig~\ref{fig:tradeoff}.  
In Fig~\ref{fig:tradeoff}(a), the color map indicates the remaining task ratio, where lower values correspond to a higher reduction in manual work. 
The minimum remaining task ratio is approximately 1\%, meaning that about 99\% of the task can be eliminated in the best threshold setting. 
Even in the worst threshold setting, the remaining task ratio is around 10\%, corresponding to a 90\% reduction. 
The white area denotes infeasible threshold combinations where the true threshold is lower than the false threshold.
In Fig~\ref{fig:tradeoff}(b), the blue line represents the precision for the positive threshold and the red line represents the inverse precision for the negative threshold.

To determine the thresholds, we aimed to simultaneously maximize both precision and inverse precision (Fig.~\ref{fig:tradeoff}(b)) while minimizing the residual human task rate (Fig.~\ref{fig:tradeoff}(a)). 
As a result, the positive and negative thresholds were set to 0.41 and 0.25, respectively. 
At these thresholds, the residual task rate is reduced to 0.63\%, corresponding to a task reduction of 99.37\%.
At this point, the misidentification rate for images with objects is 0.08\%, while the misdetection rate for images without objects reaches 5\%.

\begin{figure}[htbp]
  \centering
  \includegraphics[width=\linewidth]{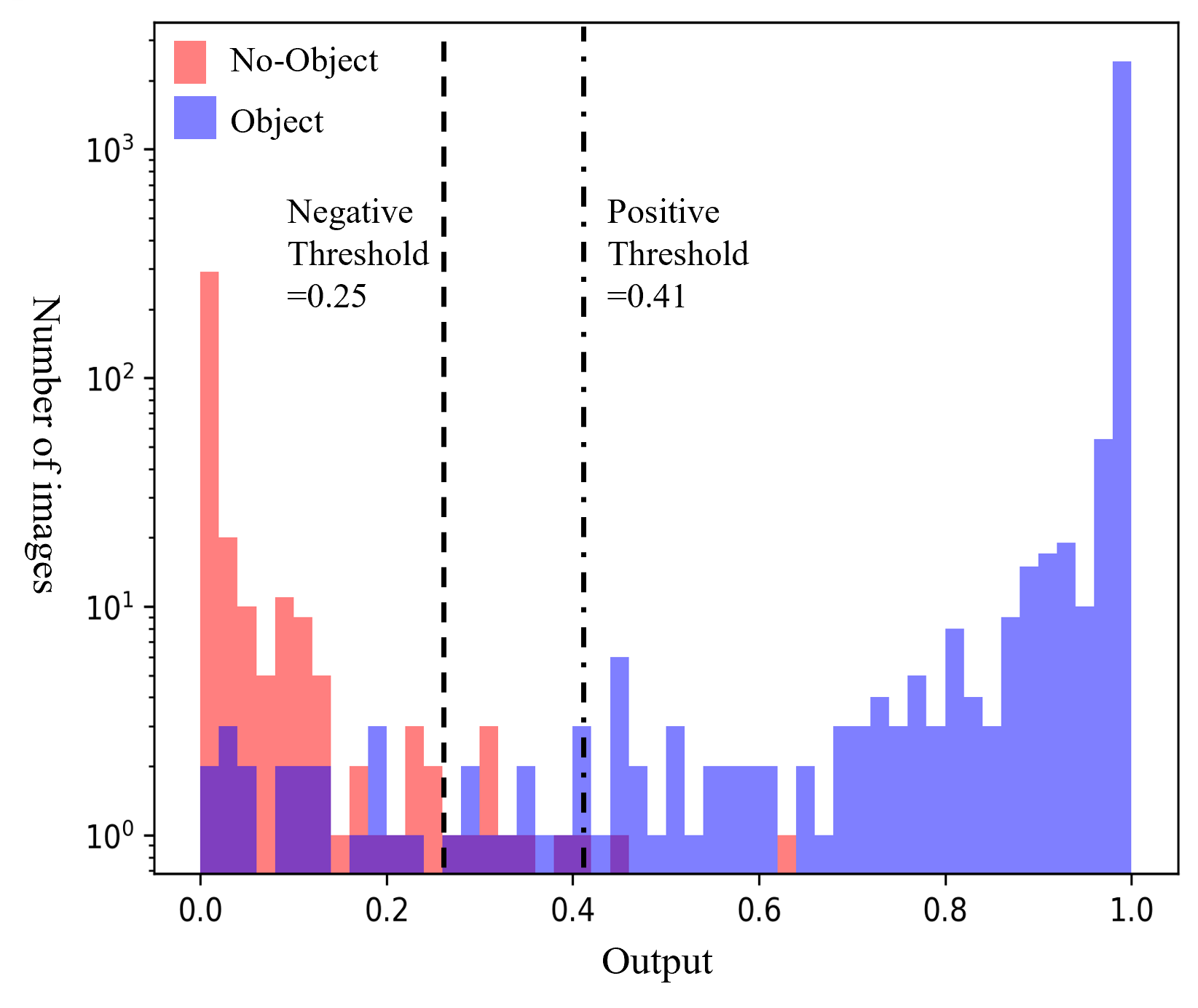}
  \caption{The histogram of the CNN3 output. The x-axis represents the probability of the object's existence, and the $y$-axis represents the number of corresponding images. The blue bars represent actual object images, and the red bars correspond to false positive images. The black line shows the threshold determined by the trade-off. {Alt text: Histogram of predicted probabilities with separate bars for objects and false positives; a vertical line marks the selected decision threshold that partitions automatic decisions from cases requiring manual review.}}
  \label{fig:hist_cnn3}
\end{figure}

The histogram in figure~\ref{fig:hist_cnn3} shows the classification results of the test dataset for model CNN3.  
The $x$-axis represents the probability of the object's existence, and the $y$-axis represents the number of corresponding images.  
The blue bars represent actual object images, and the red bars correspond to false positive images.  
The black line shows the threshold determined by the trade-off.  
Areas outside the thresholds are automatically judged by the model, while areas inside the thresholds are reviewed by humans.  
With 0.63\% of images falling within these inner regions, the model reduces the manual verification workload by 99.37\%.  

\clearpage

\section{Summary and Discussion}\label{sec:discussion}
\subsection{Summary\label{ssec:summary}}
We summarize the key outcomes of this study as follows:
\begin{itemize}
\item
We have developed a deep learning-based object detection model that incorporates the Convolutional Block Attention Module (CBAM) to process stacked astronomical images.
Our primary purpose is to examine the authenticity of the object images that the JAXA system generates.
\item
Our model is designed to extract discriminative features from overlapping multi-resolution inputs and to accurately detect moving objects from output images from the JAXA system.
\item
Evaluation on a dataset of approximately 2,000 Subaru HSC images demonstrated the model’s high performance, achieving an accuracy of 98.87\%, an F1 score of 99.35\%, and an AUC of 0.9924.  
\item 
By adjusting the detection threshold, the model reduced the human verification workload by 99.37\%, underscoring its effectiveness in replacing manual vetting in large-scale surveys.
\item 
The best-performing configuration employed a 4-layer CNN with 32–32–64–64 channels(named CNN3), using a combined input of 4- and 32-stacked images. Details of the training setup, including the optimization parameters and data augmentation methods, are summarized in Table~\ref{tab:exp_setup}.
\end{itemize}

\subsection{Discussion : Strategies for enhanced classification performance\label{ssec:performance-enhancement}}
The CBAM-enhanced CNN model demonstrated strong capability in detecting moving celestial objects.
However, several challenges remain, particularly in improving classification reliability and ensuring seamless integration into scalable operational pipelines.
In practical deployment, reducing the burden of manual vetting remains a critical goal.
As shown in Figure~\ref{fig:hist_cnn3}, overlapping label regions in the color bar (e.g., purple areas) highlight ambiguous cases that still require human inspection, despite the system achieving a 97\% reduction in verification workload. 
Enhancing the model’s ability to distinguish such borderline cases would contribute to further improvements in AUC, thereby enhancing overall screening efficiency.
To address these issues and further improve both accuracy and usability, we propose four key directions as follows (Sections
\ref{sssec:imbalance-solutions},
\ref{sssec:parameter-optimization},
\ref{sssec:learning-paradigms},
\ref{sssec:architecture-optimization},
\ref{sssec:data-strategies}
).

\subsubsection{Addressing data imbalance: immediate solutions\label{sssec:imbalance-solutions}}
Although the model was trained in a supervised fashion, the dataset is significantly imbalanced, containing far more positive samples (images with objects) than negative ones (Table~\ref{tab:ml_datasets}). This class imbalance in training datasets has been recognized as a critical challenge in ML applications for astronomical object detection.
Sample weighting techniques assign larger weights to minority class data points and smaller weights to majority class data points, encouraging models to prioritize learning from the minority class. The ATLAS asteroid survey \citep{Rabeendran2021} employed a weighted loss scheme that assigned different weights to real and false categories within the mean squared error loss function. The ZTF comet detection project \citep{Duev2021} demonstrated the effectiveness of sample weighting in highly imbalanced conditions where cutouts without comets outnumbered those with comets by a factor of five.
Fraser \citep{FRASER2025229} developed a method that augments false images to adjust the ratio of false to true images to 1.6:1, subsequently incorporating this ratio into the weighting scheme. To avoid bias from brightness variations in astronomical images, the inverse of luminosity values within images was used as weights. This approach achieved precision of 98.7\% and recall of 99.4\%. However, increasing the ratio of false images beyond this threshold resulted in a significant increase in false positive rates, indicating the importance of optimal ratio selection.

The applicability to the present study differs from conventional research contexts. Previous studies aimed to suppress false positive rates (misclassifying false images as astronomical objects) in situations where false images outnumbered astronomical images. In contrast, the present study demonstrates precision values approaching 100\% even when false images are fewer than astronomical images (Figure~\ref{fig:exp2_pre}). However, the classification accuracy for the more abundant astronomical images, corresponding to inverse precision, shows relatively low performance (Table~\ref{fig:exp2_invpre}). Therefore, strategies to address training data bias should focus on improving classification accuracy for astronomical images. The brightness-based sample weighting approach proposed by Fraser \citep{FRASER2025229} could potentially reduce false negative rates in this context.

\subsubsection{Optimization of training parameters\label{sssec:parameter-optimization}}
All hyperparameters used in this study were determined through grid search, which, while systematic, may not identify optimal parameter combinations efficiently. More sophisticated hyperparameter optimization frameworks such as Optuna \citep{akiba2019optuna} can be employed to tune learning rates, regularization coefficients, and other parameters more effectively. These frameworks utilize advanced search algorithms such as Tree-structured Parzen Estimator (TPE) and Bayesian optimization, which can significantly reduce the number of training iterations required while potentially discovering better parameter configurations. This approach could enhance both training efficiency and model performance with minimal implementation effort.

\subsubsection{Learning paradigm alternatives\label{sssec:learning-paradigms}}
The class imbalance suggests the task could be reframed as an anomaly detection problem. Specifically, unsupervised learning could be used to detect object-free images as anomalies, potentially circumventing the challenges associated with imbalanced supervised learning.
Contrastive learning methods such as SimCLR \citep{chen2020simclr} and MoCo \citep{he2020moco} have proven effective in learning robust feature spaces from limited data. Both are categorized as unsupervised approaches that could leverage the abundant unlabeled astronomical image data. Semi-supervised learning methods such as FixMatch \citep{sohn2020fixmatch} and Mean Teacher \citep{tarvainen2018meanteachers} are also promising for achieving high performance with minimal labeled data. Hybrid models like SelfMatch \citep{kim2021selfmatch} can further leverage both labeled and unlabeled data, balancing generalization and accuracy.

\subsubsection{Model architecture optimization\label{sssec:architecture-optimization}}
While the current ensemble uses multiple models of identical architecture with different initializations, future work could explore heterogeneous ensembles across diverse architectures, such as ResNet \citep{he2015} and EfficientNet \citep{tan2020efficientnet}. Investigating depth and width variations within the same family could also reveal architectures more resilient to noise or overfitting. Architectural diversity has been shown to mitigate model-specific biases and improve both average accuracy and prediction stability.

A promising direction is to treat multiple image stacks with the same number of frames (e.g., four 8-frame stacks) as time-series data. This opens the door to spatiotemporal modeling using architectures such as 3D Convolutional Neural Networks (3D-CNNs) \citep{Ji2013_3D_CNN} and Convolutional Long Short-Term Memory networks (ConvLSTMs) \citep{shi2015}, which can learn temporal differences between moving and stationary celestial objects. 

\subsubsection{Advanced data strategies\label{sssec:data-strategies}}
Our experimental results revealed that CNNs trained on only 32-frame stacks exhibited high variance and lower performance, while CNNs trained on multi-depth stacks (e.g., 32+16, 32+8) performed better and more consistently (Figure~\ref{fig:exp1_acc}-\ref{fig:exp1_f1}). This discrepancy may be due to the mismatch between pretrained ImageNet\cite{Deng2009} weights and the structure of stacked images—where each channel contains a different frame, unlike RGB images where all channels depict the same content. The pretrained weights may thus degrade performance in this context.

Training from scratch using larger datasets would alleviate this issue. However, standard augmentations such as rotation and flipping are insufficient for generating diverse and representative samples. Thus, synthetic data generation using physics-based simulations or Generative Adversarial Networks (GANs) \citep{goodfellow2014gan} becomes crucial. Future data augmentation could include GAN-generated background artifacts such as galaxies and nebulae, or physics-based noise injection to simulate realistic observational conditions. Physics-Informed Neural Networks (PINNs) \citep{Rassi2019} could further support physically consistent augmentation by incorporating underlying astrophysical models.

Beyond image data, future models may integrate multimodal inputs including light curves, spectral data, and ephemeris information. Such inputs would support not only object detection but also downstream inference of physical characteristics such as size, brightness variation, and orbital elements.

\subsection{Discussion : Integration into operational pipelines and broader applications\label{ssec:summary-pipeline}}
We plan to deploy our detection model into the final processing pipeline of the JAXA survey system. This integration will enable fully automated detection of faint moving objects and improve the operational efficiency and scientific reliability of future missions, including Kuiper Belt Object (KBO) flybys.
The methodology developed here can be extended to future large-scale surveys such as the Vera C. Rubin Observatory’s Legacy Survey of Space and Time (LSST) \citep{ivezic2019} and the Square Kilometre Array (SKA) \citep{braun2019}. These projects are expected to generate unprecedentedly large and complex datasets, where traditional analysis methods alone will not suffice. ML-based approaches like ours will be essential for scalable, precise, and automated detection of faint moving objects.

\begin{ack}
This research is based on data collected at the Subaru Telescope operated by the National Astronomical Observatory of Japan (NAOJ), and the data is obtained from SMOKA operated by Astronomy Data Center (ADC), NAOJ.
We are honored and grateful for the opportunity of observing the Universe from Maunakea, which has the cultural, historical, and natural significance in Hawaii.
The observations were supported in part by NASA for the New Horizons mission.
In particular, NASA Keck exchange time was used to acquire the data on 17 June 2021 (S21A--TE216--K).
Numerical computations were carried out on GPU cluster at the Center for Computational Astrophysics, National Astronomical Observatory of Japan.
F.~Y. acknowledges support from the Japan Society for the Promotion of Science (JSPS) Kakenhi grant, 24K07123.
J.~J. acknowledges support from the Japan Society for the Promotion of Science (JSPS) Kakenhi grants, JP19K23456 and JP22K14069.
A.~V. acknowledges support from JPL/RSA contract no. 1659537.
The authors used Overleaf to provide a collaborative and efficient online {\LaTeX} environment, which facilitated the preparation and formatting of this manuscript.
This study has made use of NASA's Astrophysics Data System (ADS) Bibliographic Services
We thank Alan Stern for his detailed feedback which significantly improved the presentation and quality of the article..
\end{ack}

\bibliography{shibukawa}{}
\bibliographystyle{aasjournal} 

\end{document}